\documentclass[11pt]{article}

\usepackage[final]{acl}

\usepackage{times}
\usepackage{latexsym}

\usepackage{pifont}
\usepackage{fontawesome5} 
\usepackage{booktabs} 

\usepackage[T1]{fontenc}

\usepackage[utf8]{inputenc}

\usepackage{microtype}

\usepackage{inconsolata}

\usepackage{graphicx}
\usepackage{multirow}%
\usepackage{makecell}
\usepackage{colortbl}
\usepackage{tabularx}
\usepackage{amssymb}
\usepackage{subfig}

\usepackage{xcolor} 
\definecolor{first}{RGB}{153,204,255}   
\definecolor{second}{RGB}{191,219,255}  
\definecolor{third}{RGB}{224,236,255}   
\definecolor{lightred}{RGB}{255, 180, 180}    
\definecolor{lightgreen}{RGB}{180, 255, 180}  
\title{EchoVLM: Dynamic Mixture-of-Experts Vision-Language Model for Universal Ultrasound Intelligence}

\author{
	\textbf{Chaoyin She\textsuperscript{1}},
	\textbf{Ruifang Lu\textsuperscript{2}},
	\textbf{Lida Chen\textsuperscript{2,*}},
	\textbf{Wei Wang\textsuperscript{2,*}},
	\textbf{Qinghua Huang\textsuperscript{1,3,*}}
	\\
	\textsuperscript{1}School of iOPEN, Northwestern Polytechnical University, Xi'an, China\\
	\textsuperscript{2}The First Affiliated Hospital of Sun Yat-Sen University, Guangzhou, China\\
	\textsuperscript{3}College of Mechanical Engineering, Tongji University, Shanghai, China
	\\
	\small{
		\texttt{\{chenlda, wangw73\}@mail.sysu.edu.cn} \quad
		\texttt{qinghua\_huang@tongji.edu.cn} \quad
		\texttt{qhhuang@nwpu.edu.cn}
	}
}

\begin{document}
\maketitle
\begin{abstract}
Ultrasound is the preferred early cancer screening modality due to non-ionizing radiation, cost-effectiveness, and real-time imaging, yet conventional diagnosis relies heavily on physician expertise, causing significant subjectivity and limited efficiency. Vision-Language Models (VLMs) show promise but lack ultrasound-specific knowledge and multi-organ generalization. We propose EchoVLM, the first open-source 10-billion-parameter ultrasound-tailored VLM with a Mixture-of-Experts (MoE) architecture. It is infused with knowledge across seven anatomical systems, trained on 208,941 clinical cases, 1.47 million ultrasound key-frame images, and over 100 diseases or imaging findings. Supporting clinical report generation, diagnosis prediction, and Visual Question Answering (VQA), it outperforms Qwen2-VL by 7.58 BLEU-1 and 3.45 ROUGE-1 points in report generation. This work shows substantial potential for establishing a general-purpose ultrasound VLM and lays a technical foundation for clinical translation. Source code and model weights are available at \url{https://github.com/Asunatan/EchoVLM}.
\begingroup
\renewcommand{\thefootnote}{\fnsymbol{footnote}}
\footnotetext[1]{Corresponding authors
}
\endgroup

\end{abstract}
\section{Introduction}
Ultrasound imaging has become a cornerstone of clinical diagnostics, distinguished by the absence of ionizing radiation, cost-effectiveness, and real-time dynamic visualization capabilities—attributes that render it indispensable for early cancer detection, prenatal care, and dynamic assessment of organ structure and function. However, conventional ultrasound diagnosis remains heavily dependent on radiologists’ specialized expertise, with manual interpretation introducing inter-observer variability, diagnostic delays, and suboptimal treatment efficiency.
\begin{figure}[ht]
	\centering
	\includegraphics[width=\linewidth]{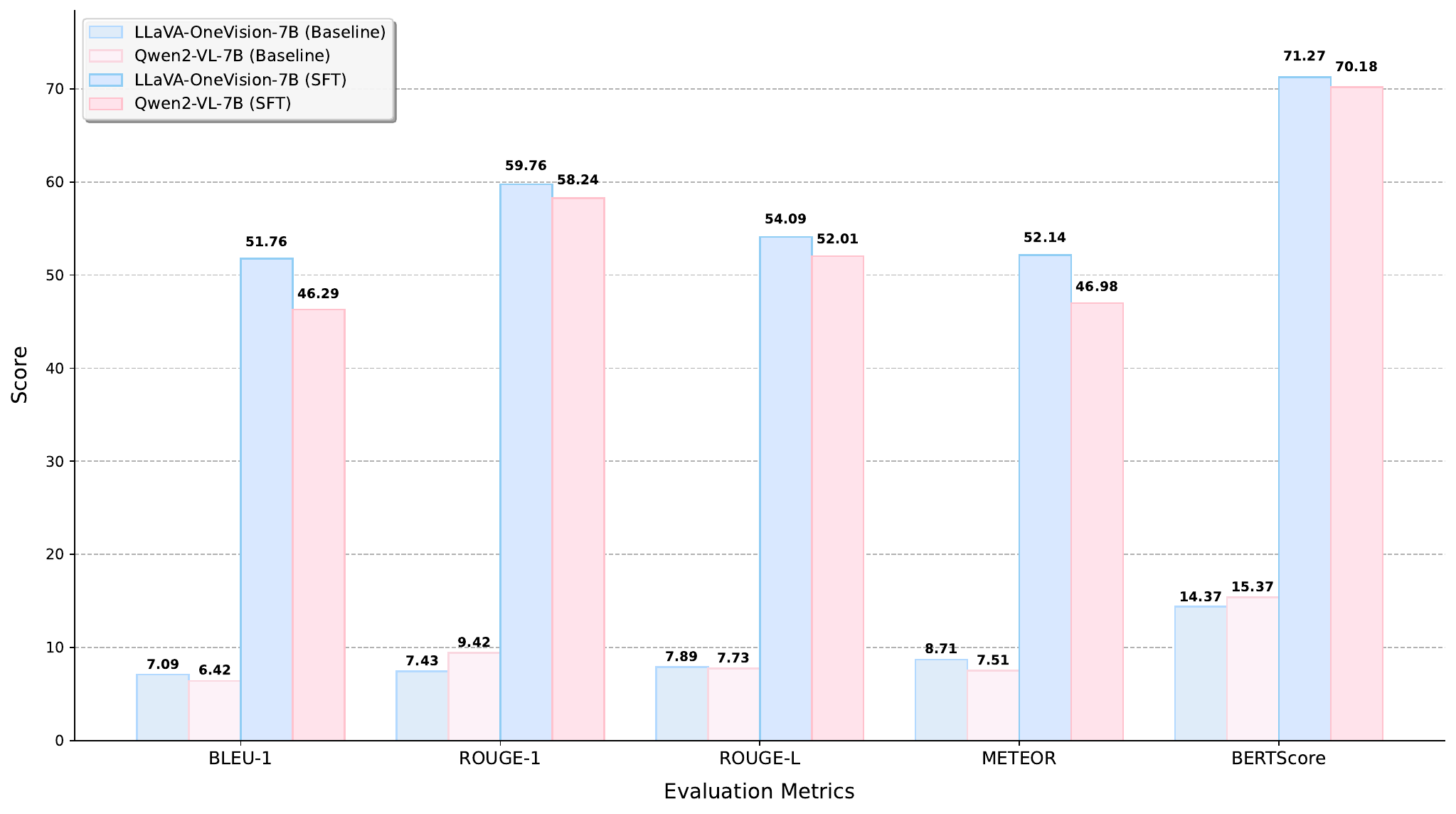} 
	\caption{Comparative performance evaluation of generic and ultrasound-specialized VLMs.}
	\label{fig_1}
\end{figure}
While Visual Language Models (VLMs) have made significant advances in multimodal perception, their application to ultrasound diagnosis faces a critical domain-specific limitation: general-purpose VLMs exhibit limited domain generalization to ultrasound medicine. Our preliminary experiments (Figure ~\ref{fig_1}) reveal a substantial performance gap between baseline generic models and their supervised fine-tuned (SFT) variants, confirming that off-the-shelf VLMs cannot effectively capture the nuanced, domain-specific features of ultrasound images. This limitation directly restricts their clinical utility in ultrasound workflows, underscoring the urgent need to develop ultrasound-specialized VLMs.

To address this gap, we introduce EchoVLM, the first universal ultrasound-specialized VLM with tens of billions of parameters, founded on three core innovations: (1) We curated the largest multi-organ ultrasound dataset to date, covering seven anatomical systems based on 208,941 clinical cases from 15 hospitals, 1.47 million key-frame ultrasound images, and over 100 related diseases and imaging findings, ensuring comprehensive coverage for robust model training. (2) Inspired by Self-Instruct~\cite{Self-Instruct}, we propose an expert-validated few-shot prompting mechanism to build a multi-task instruction-tuning data generation pipeline, which has generated 1.8 million pairs of instruction-tuning data by synthesizing diverse diagnostic scenarios while ensuring clinical accuracy via expert oversight. (3) We utilize a Dual-path MoE module for knowledge injection while preserving pre-acquired knowledge; its dynamic routing mechanism enhances adaptability to ultrasound’s heterogeneous complexity, enabling task-specific subnetworks to specialize in distinct diagnostic domains for improved efficiency and precision.
In summary, our key contributions are:
\begin{itemize}
	\item We pioneer EchoVLM, the first universal ultrasound-specialized VLM with tens of billions of parameters tailored to address ultrasound diagnostic challenges.
	\item We curate a large-scale multi-organ ultrasound dataset (208,941 cases from 15 hospitals, 1.47 million key-frame images, over 100 diseases/imaging findings, covering seven anatomical systems) and develop an expert-validated multi-task instruction-tuning pipeline generating 1.8 million data pairs.
	\item We integrate a Dual-path MoE module into EchoVLM for knowledge injection while preserving pre-acquired knowledge, whose dynamic routing mechanism markedly enhances adaptability to ultrasound task complexity. 
\end{itemize}
\begin{table*}[h]
	\centering
	\resizebox{1.0\linewidth}{!}{%
		\footnotesize 
		\begin{tabular}{lp{2.0cm}p{2.8cm}p{3.5cm}p{3.5cm}c}
			\hline
			Dataset & \multicolumn{1}{l}{\makecell{Anatomical \\ Coverage}} & \multicolumn{1}{l}{\makecell{Dataset Scale}} & \multicolumn{1}{l}{Imaging Modality} & \multicolumn{1}{l}{Target Tasks} & \multicolumn{1}{l}{Multimodal} \\
			\hline
			BUSI ~\cite{BUSI} & Breast &  \makecell[l]{780 images\\ 600 patients} & \makecell[l]{2D B-mode} & \makecell[l]{Segmentation \\ Classification} & \textcolor{red}{\faTimes} \\
			\hline
			OASBUD ~\cite{OASBUD} & Breast & \makecell[l]{200 images\\ 100 patients} & \makecell[l]{RF Ultrasound \\ 2D B-mode} & \makecell[l]{Segmentation \\ Classification \\ QUS Analysis} & \textcolor{red}{\faTimes} \\
			\hline
			BUS-UCLM ~\cite{BUS-UCLM} & Breast & \makecell[l]{683 images \\ 38 patients} & \makecell[l]{2D B-mode} & \makecell[l]{Segmentation \\ Classification} & \textcolor{red}{\faTimes} \\
			\hline
			TN3K ~\cite{TN3K} & Thyroid & \makecell[l]{3,493 images \\ 2,421 patients} & \makecell[l]{2D B-mode} & Segmentation & \textcolor{red}{\faTimes} \\
			\hline
			TNUI-2021 ~\cite{N-Net} & Thyroid & \makecell[l]{1,381 images \\ 483 patients} & \makecell[l]{2D B-mode} & \makecell[l]{Segmentation \\ Classification} & \textcolor{red}{\faTimes} \\
			\hline
			DDTI ~\cite{DDTI} & Thyroid & \makecell[l]{134 images \\ 99 patients} & \makecell[l]{2D B-mode} & \makecell[l]{Segmentation \\ Classification} & \textcolor{red}{\faTimes} \\
			\hline
			EchoNet-Dynamic \cite{EchoNet-Dynamic} & Heart & \makecell[l]{10,030 videos \\  10,030 patients} & \makecell[l]{Echocardiography Videos \\ Color Doppler \\ Spectral Doppler \\ Tissue Doppler} & \makecell[l]{Segmentation, \\ Function Assessment} & \textcolor{red}{\faTimes} \\
			\hline
			EchoPrime \cite{EchoPrime} & Heart & \makecell[l]{12.1M videos \\ 275K studies \\ 67.8M text tokens} & \makecell[l]{Echocardiography Videos \\ Color Doppler \\ Spectral Doppler \\ Tissue Doppler}& \makecell[l]{VLM Pre-training\\Classification \\ Diagnosis\\Cross-modal Retrieval\\} & \textcolor{green}{\faCheck} \\
			\hline
			KMVE \cite{li2024ultrasound} & \makecell[l]{Breast \\ Thyroid \\ Liver} & \makecell[l]{7,390 patients \\ 7,390 reports} & \makecell[l]{2D B-mode} & \makecell[l]{VLM Pre-training\\Image Captioning
				 \\  Report Generation\\Description Generation} & \textcolor{green}{\faCheck} \\
			\hline
			FetalCLIP \cite{FetalCLIP} & Obstetrics & \makecell[l]{210,035 images \\ Paired clinical text} & \makecell[l]{2D B-mode \\ Color Doppler} & \makecell[l]{VLM Pre-training \\Classification\\Segmentation\\Detection } & \textcolor{green}{\faCheck} \\
			\hline
			EchoCLIP \cite{EchoCLIP} & Heart & \makecell[l]{1.03M videos\\99,870 patients \\ 99,870 clinical report} & \makecell[l]{Echocardiography Videos \\ Color Doppler \\ Spectral Doppler \\ Tissue Doppler} & \makecell[l]{VLM Pre-training\\Classification\\ Regression\\Cross-modal Retrieval} & \textcolor{green}{\faCheck} \\
			\hline
			Sonomate \cite{Sonomate} & Obstetrics & \makecell[l]{525 unique video \\and  audio pairs\\2.7M frames \\ 63.8K sentences} & \makecell[l]{Fetal Ultrasound \\ Videos} & \makecell[l]{VLM Pre-training \\ Classification\\Detection\\Cross-modal Retrieval\\VQA} & \textcolor{green}{\faCheck} \\
			\hline
			EchoVLM (Ours) & \makecell[l]{7 Organ \\ Systems} & \makecell[l]{1.47M images \\ 208K patients \\ 1.8M instruction pairs} & \makecell[l]{2D B-mode \\Color Doppler\\Spectral Doppler\\Tissue Doppler } & \makecell[l]{Report Generation \\ VQA\\ Diagnosis \\ } & \textcolor{green}{\faCheck} \\
			\hline
		\end{tabular}%
	}
	\caption{Summary of Medical Ultrasound Datasets.}
	\label{tab:us_datasets}
\end{table*}
\section{Related Work}
Large language models (LLMs) have driven paradigm shifts in artificial intelligence, especially in natural language processing (NLP) for comprehension, generation, and text-based tasks. However, traditional unimodal LLMs inherently lack visual perception capabilities, which limits their applicability to real-world cross-modal scenarios. To address this, researchers first leveraged image-text alignment methods~\cite{clip,MobileCLIP,LongCLIP} to establish a foundational bridge between visual and textual modalities; subsequently, semantic embedding layers~\cite{llava,VideoLLaVA,VILA} and cross-attention layers~\cite{BLIP2,Flamingo} have been proposed to achieve effective visual-textual feature integration. Notably, the BLIP series pioneered cross-attention for dynamic visual-textual interaction, while LLaVA introduced visual instruction tuning to enhance perceptual capabilities via visual data fine-tuning. Subsequent advancements improved performance through dataset expansion~\cite{LRVInstruction,LLaVAR}, higher image resolution~\cite{LLaVA1_5,CogVLM,CogVLM2}, multi-image/video understanding~\cite{LLaVAOneVision,LLaVAvideo}, projection layer optimization~\cite{MMFuser,rethinkingvisuallayerselection,Eyes}, and multi-visual encoders~\cite{BRAVE,MouSi}. Recent progress focuses on fine-grained understanding by integrating object localization (e.g., bounding boxes) and segmentation masks to boost spatial and semantic accuracy~\cite{RegionGPT,omgllava}.
Notably, VLMs have been increasingly applied in medicine, with LLaVA-Med~\cite{LLaVAMed}, Medgemma~\cite{Medgemma}, HuatuoGPT-Vision~\cite{HuatuoGPT-Vision}, and Lingshu~\cite{Lingshu} as representative examples. However, such systems face critical limitations in specialized clinical settings, especially in terms of insufficient ultrasound-specific knowledge that impairs structured report generation. Additionally, their limited context length fails to handle the multi-image scenarios common in ultrasound. To address these domain-specific issues, this study proposes an ultrasound-specific VLM architecture optimized for diagnostic workflows and standardized sonographic terminology.

\section{Method}
\subsection{Data Collection and Instruction-Tuning Data Generation Pipeline}
\begin{figure*}[ht]
	\centering
	\includegraphics[width=\linewidth]{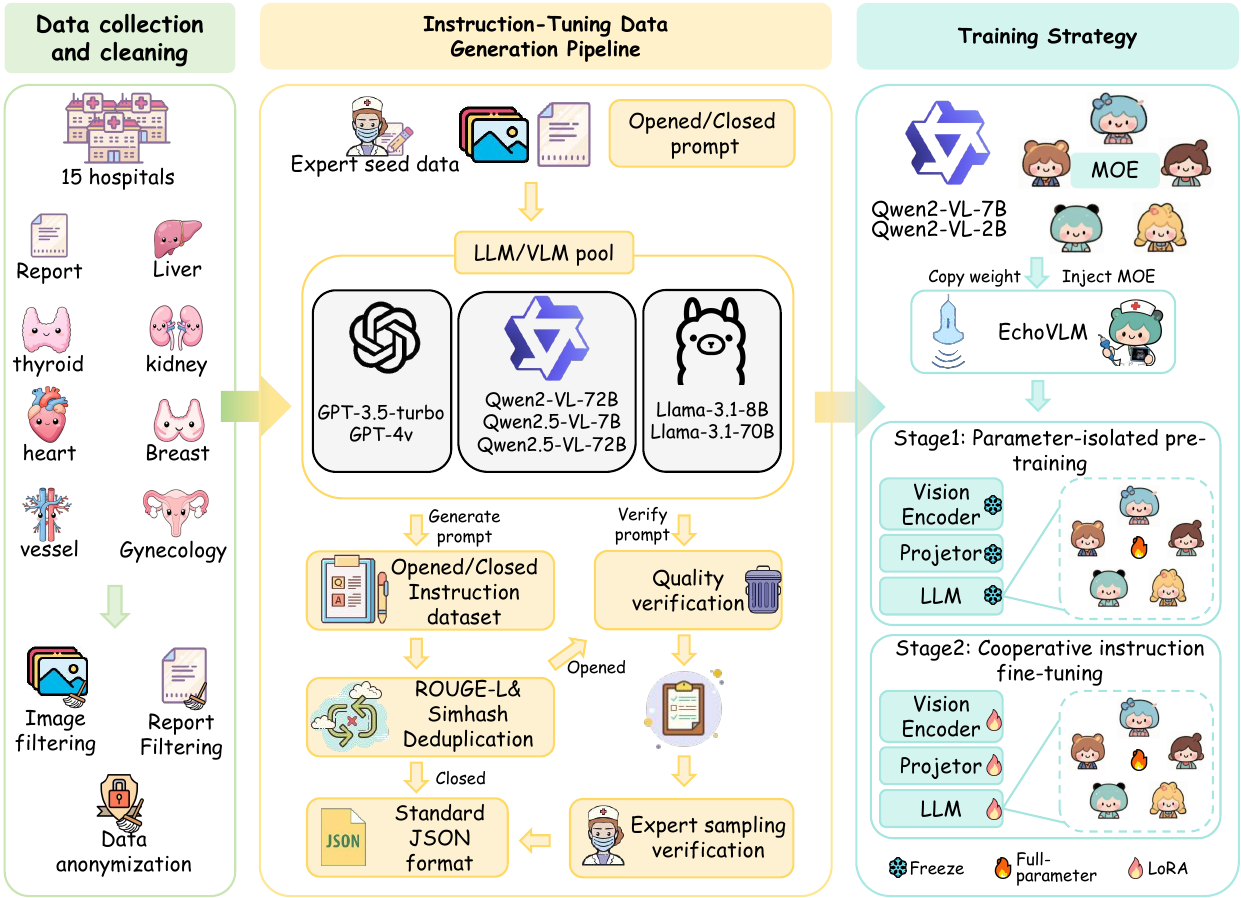}
	\caption{Overview of EchoVLM framework.}
	\label{fig_2}
\end{figure*}

\begin{figure*}[ht]
	\subfloat[Training dataset distribution]{
	\includegraphics[width=0.48\linewidth]{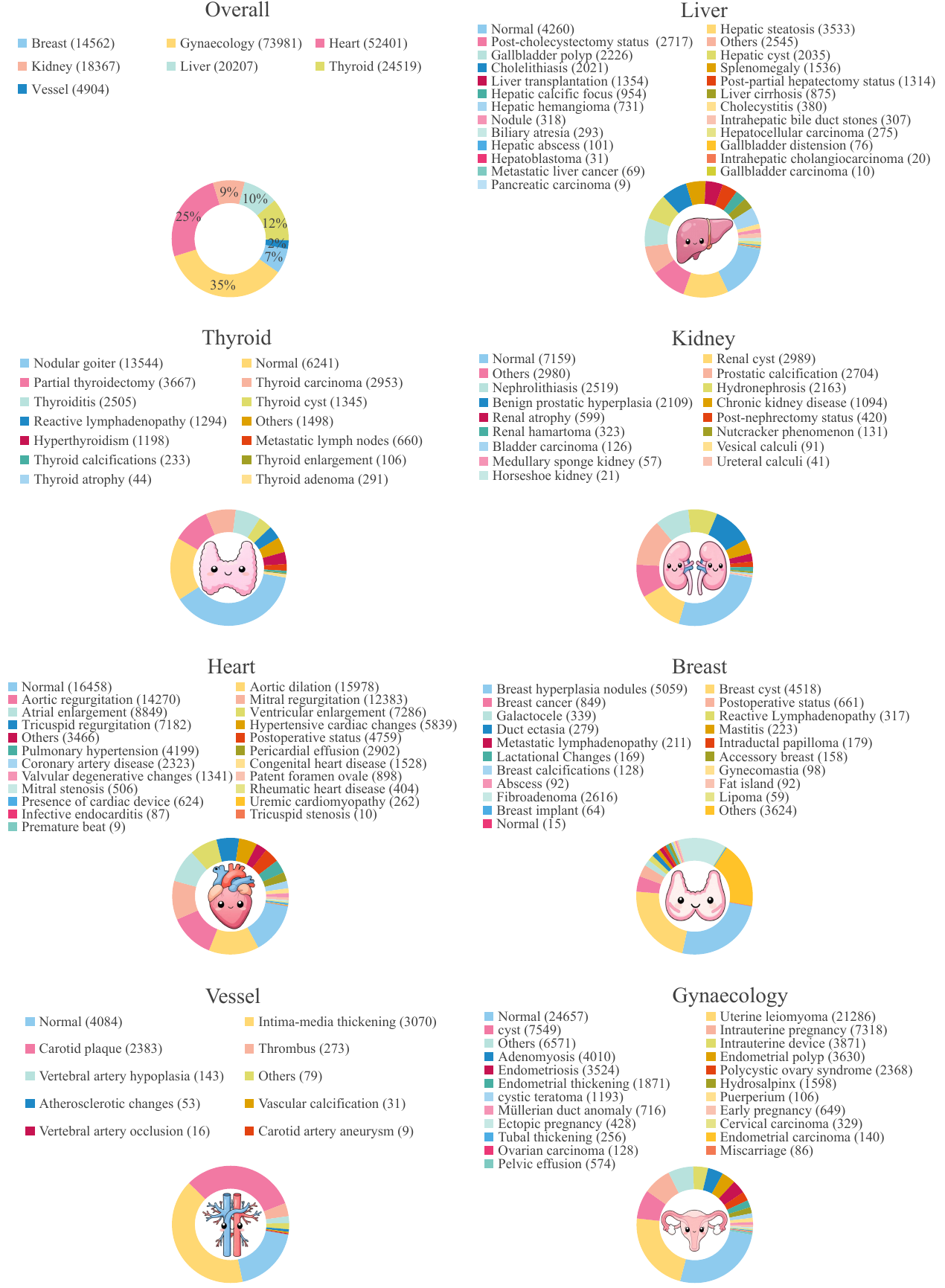}} \hfill
	\subfloat[Testing dataset distribution]{
		\includegraphics[width=0.48\linewidth]{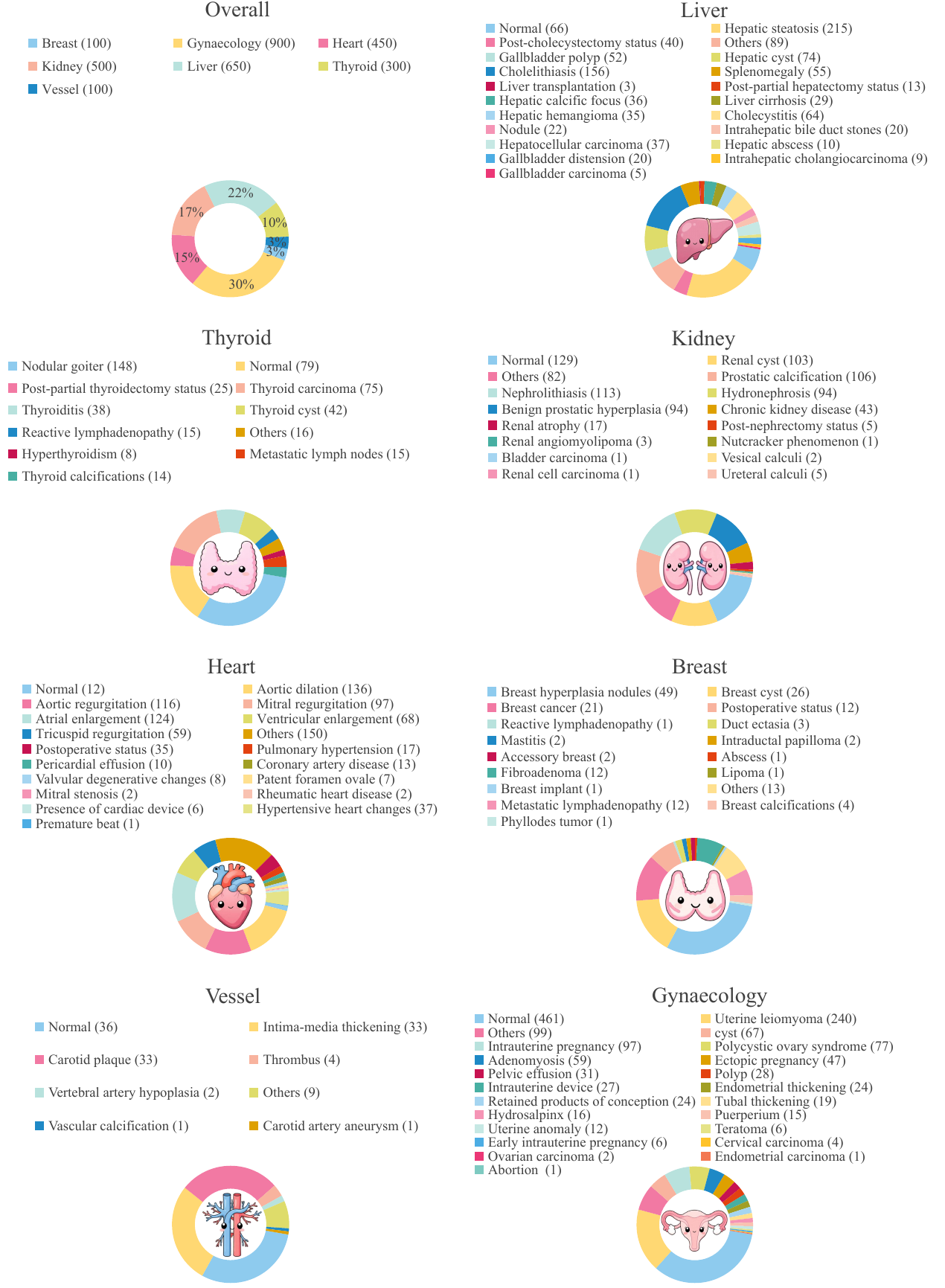}}

	\caption {Distribution of cases across seven anatomical systems in a multicenter ultrasound dataset, with numbers in parentheses indicating case counts. Note that a single case may involve multiple diseases.}
	\label{fig_3}
\end{figure*}

To develop a VLM tailored to ultrasound imaging, we compiled a comprehensive dataset from 15 hospitals, as shown in Figure~\ref{fig_2}. It covers seven major anatomical systems commonly assessed via ultrasound: liver, kidneys, thyroid, vascular system, gynecological organs, heart, and breasts.
A rigorous data filtration protocol ensured quality: (1) Image Filtering: Only single-region images were extracted from hospital databases to avoid multi-region ambiguity. Images without corresponding reports were manually removed for alignment between imaging and clinical data. (2) Text Filtering: Sensitive patient information was deleted using regular expression matching to ensure data privacy, and irrelevant reports or those lacking imaging data were manually deleted. To advance ultrasound-specific VLMs, we pioneered the redefinition of data desensitization (see Appendix~\ref{sec:appendix_a} for details). This process yielded 208,941 cases with 1.47 million key-frame images, covering over 100 diseases and imaging findings (Figure~\ref{fig_3}).

Leveraging this dataset, we established a structured instruction-tuning data generation pipeline using few-shot prompting (Figure~\ref{fig_2}). Medical experts developed 21 exemplary templates across diverse pathologies, each integrating three clinically simulative components: (1) Ultrasound reports: Detailed records of lesion location, size, morphology, echogenicity, and other image-derived features. (2) Ultrasound diagnosis: Standardized diagnostic summaries synthesizing key findings. (3) VQA pairs: Multidimensional sets covering interpretation, risk stratification, counseling, surveillance, and treatment planning.
Templates were categorized into open- and closed-ended types with tailored prompting. For open-ended ones, models generated both questions and answers via example reference; for closed-ended ones, questions were model-generated and answers extracted from real reports to ensure validity. ROUGE-L and Simhash algorithms deduplicated data. The open-ended subset underwent dual accuracy validation: automated evaluation via a VLM/LLM pool and expert manual review of random samples. This pipeline produced 1.8 million high-quality instruction-tuning pairs (see Appendix~\ref{sec:appendix_a} for details). 

To further contextualize the contributions of our proposed dataset and instruction-tuning pipeline, we conducted a rough statistical summary of existing studies in the ultrasound imaging domain, which are summarized in Table ~\ref{tab:us_datasets}). This analysis demonstrates a clear paradigm shift in the field, wherein research is rapidly transitioning from traditional single-modality, task-specific approaches to large-scale, multimodal corpora that are tailored for vision-language pretraining and multi-task learning. Such an evolutionary trend further underscores the growing necessity of comprehensive, clinically aligned datasets to empower the development of next-generation ultrasound VLMs.
\subsection{Architecture of EchoVLM}
We introduce EchoVLM, a vision-language model tailored for clinical ultrasound analysis, developed via targeted domain specialization of the Qwen2-VL~\cite{Qwen2-VL} foundational model. Instead of naïvely fine-tuning the model components on ultrasound corpora, we introduce a Dual-path MoE mechanism which injects domain knowledge while avoiding destructive updates to pre-existing representations.

For the visual encoder, any RGB ultrasound frame $ v \in \mathbb{R}^{H \times W \times 3} $ are encoded by a native dynamic resolution Vision Transformer (ViT) to generate a discrete visual token representations $ V = [v_1, v_2, \dots, v_m] \in \mathbb{R}^{M \times C} $. Here, $ M = HW / 14^2 $ denotes the number of visual tokens derived from the original spatial resolution $H\times W$. These visual tokens are subsequently transformed into the dimensional space of the Qwen-2 via a Multilayer Perceptron (MLP) projector $ f(\cdot) $, yielding dimensionally-mapped visual embeddings $ V \in \mathbb{R}^{M \times D} $. Simultaneously, textual inputs undergo parallel processing: the input prompts are first tokenized and then embedded through the word-embedding layer $ w(\cdot) $, generating the textual token sequence $ T = [t_1, t_2, \dots, t_N] \in \mathbb{R}^{N \times D} $, where $ N $ represents the number of text tokens determined by the input prompt length. The concatenated representation $ X_{0}  = [V; T] \in \mathbb{R}^{(M+N) \times D} $ is subsequently fed into the LLM which comprises multiple Transformer blocks. Each block integrates multi-head self-attention (MSA), RMS normalization (RMSNorm), residual connections, and the Dual-path MoE block, with its processing workflow formulated as follows:
\begin{equation}
	X_{i}' = MSA(RMSNorm(X_{i-1})) + X_{i-1}
\end{equation}
\begin{equation}
	X_{i} = MoE(RMSNorm(X_{i}')) + X_{i}'
\end{equation}
	\subsection{Dual-path MoE}
	Structurally, the Dual-path MoE layer comprises two complementary sets of experts.
	First, a static expert is instantiated by copying the original Qwen2 Feed-Forward Network (FFN) and immediately frozen; its parameters therefore act as a resilient anchor that conserves generic semantic capacity. Second, a battery of active experts is appended and trained. Within this group we further distinguish (1) a shared expert ($S$) that processes every token, thereby sustaining a universal ultrasound representation, and (2) a cohort of routing experts ($E$) that are sparsely activated via a top-2 gating function conditioned on token-level features.
	\begin{equation}
		\resizebox{\linewidth}{!}{
			$Y = \alpha FFN(X) + (1 - \alpha) \left[ \lambda S(X) +\sum_{i=1}^k g_{i}(X)E_{i}(TopK(X)) \right]$
		}
	\end{equation}
	\begin{equation}
		g_{i}(X) = \frac{e^{f_i(X)}}{\sum_{j=1}^k e^{f_j(X)}}
	\end{equation}
	In these equations, $g_i$ quantifies the contribution of expert $E_i$, $f_i$ denotes the routing logits, and $\alpha $ is a learnable scalar that modulates the equilibrium between generic knowledge and ultrasound-specific information.
	
	\subsection{Training Strategy}
	\subsubsection{Stage I: Parameter-isolated pre-training.}
	All original parameters of the Qwen2-VL are completely frozen, and only the newly-introduced MoE blocks are activated. This isolation strategy prevents catastrophic forgetting of generic multimodal knowledge, steering the expert layers to exclusively acquire domain-specific ultrasound representations.
	\subsubsection{Stage II: Cooperative instruction fine-tuning.}
	We introduce a collaborative optimization mechanism by applying low-rank adaptation (LoRA:$
	\mathbf{W} = \mathbf{W}_0 + \Delta \mathbf{W}$, where $\Delta \mathbf{W} = \mathbf{A} \mathbf{B}^\top$ with $\mathbf{A} \in \mathbb{R}^{d \times r}$, $\mathbf{B} \in \mathbb{R}^{d \times r}$) ~\cite{LoRA} for lightweight parameter tuning of the base model, while maintaining full parameter updates for the MoE components. Notably, for the architecture comprising frozen FFN layers and activated MoE modules, we develop dynamic modulation parameters $\lambda \in [0,1]$ to balance the contributions between general world knowledge and ultrasound domain knowledge. 
	\subsubsection{Training Objectives.}
	We optimize the model by minimizing a compound loss that balances the auto-regressive loss with an expert-load–balancing penalty. The vision encoder processes input images to generate $m$ visual tokens $V$. For textual inputs, raw text undergoes tokenization and is subsequently mapped to dense embeddings using an embedding layer, yielding $n$ textual tokens $T$. Their concatenation forms the model input $X=[V,T]$ of length $K=m+n$. During training, the model autoregressively generates a response of length $L$; the autoregressive loss is thus defined as:
	\begin{equation}
		\mathcal{L}_{ar} = - \sum_{i=1}^{L} \log p_{\theta}(y_i \mid X, y_{<i})
	\end{equation}
	where $\theta $ is a trainable parameter. To ensure balanced utilization of the $E$ experts, we introduce an auxiliary load-balancing loss.  Let $g_e(t)\in [0,1]$ denote the gating weight for expert $e$ and token $t$, where $\sum_{e=1}^{E} g_e(t) = 1$. For a mini-batch of tokens $\mathcal{B}$, we compute the dispatch ratio $F_e$, representing the fraction of tokens routed to expert $e$ :
	\begin{equation}
		F_e = \frac{1}{|\mathcal{B}|} \sum_{t \in \mathcal{B}} \mathbb{I}\left[g(t) \rightarrow e\right]
	\end{equation}
	where $g(t) \rightarrow  e$ denotes that token $t$ is routed to expert $e$, $\mathbb{I}[\cdot ]$ is the indicator function that equals 1 if token $t$ is routed to expert $e$, and 0 otherwise. The average gating probability is defined as:
	\begin{equation}
		G_e = \frac{1}{|\mathcal{B}|} \sum_{t \in \mathcal{B}} g_e(t).
	\end{equation}
	The auxiliary balancing loss is then given by:
	\begin{equation}
		\mathcal{L}_{bal} = \sum_{e=1}^{E} F_e G_e.
	\end{equation}
	The overall training objective combines both components:
	\begin{equation}
		\mathcal{L}_{total} = \mathcal{L}_{ar} + \gamma \mathcal{L}_{bal}
	\end{equation}
	where $\gamma $ is a hyperparameter controlling the strength of the load-balancing penalty.

\section{Experiments}
\begin{table*}[h]
	\small 
	\centering
	\resizebox{\linewidth}{!}{%
		\begin{tabular}{c|c|cccccccccccc}
			\hline
			Anatomical	&Metric$\uparrow$&\makecell{Medgemma\\(4B)}&\makecell{LLaVA1.5\\(7B)}&\makecell{HuatuoGPT-Vision\\(7B)}& \makecell{Lingshu\\(7B)} & \makecell{LLaVA-OneVision\\ (7B)}&\makecell{Qwen2-VL\\ (7B)}&\makecell{LLaVA-Med\\(7B)}&\makecell{Qwen2.5-VL\\(7B)}&\makecell{LLaVa-NeXT\\(13B)}&\makecell{Gemma-3\\(12B)}&\makecell{EchoVLM\\(3B)} &\makecell{EchoVLM\\(11B)} \\ 
			\hline
			\multirow{5}{*}{Breast} 
			&BLEU-1		&21.91&46.49& 50.10  & 48.99& 49.59 &41.74	&\cellcolor{first}\textbf{51.17}	&50.55&49.50&48.63&\cellcolor{third}50.55&\cellcolor{second}50.76\\ 
			&ROUGE-1 	&41.13&57.07& 60.97  & 60.87& 61.45	&\cellcolor{second}62.85	&60.44			&61.78&59.03&60.49&\cellcolor{third}62.24&\cellcolor{first}\textbf{64.38}\\
			&ROUGE-L 	&34.11&49.33& 51.64  & 52.18& 51.47	&\cellcolor{second}53.46	&52.03			&52.38&50.53&49.59&\cellcolor{third}53.31&\cellcolor{first}\textbf{55.21}\\ 
			&METEOR 	&26.42&45.29& 49.76	 & 49.32& \cellcolor{second}49.86	&47.32	&48.41			&49.51&47.11&47.44&\cellcolor{third}49.8&\cellcolor{first}\textbf{50.14}\\ 
			&BERTScore	&51.36&66.92& 70.35	 & 70.39& 70.84	&\cellcolor{second}72.21	&69.50			&\cellcolor{third}71.53&68.32&70.35&70.82&\cellcolor{first}\textbf{73.00}\\
			\hline
			\multirow{5}{*}{Gynecology} 
			&BLEU-1		&46.33&43.51& 48.75	& \cellcolor{second}51.40	& 47.83&\cellcolor{third}49.88&42.63&47.28&43.01&49.05&45.79&\cellcolor{first}\textbf{52.52}\\
			&ROUGE-1	&55.68&48.80& 53.18	& \cellcolor{third}56.64 & 54.98&\cellcolor{second}56.97&51.64&52.15&52.06&55.71&53.15&\cellcolor{first}\textbf{59.19}\\
			&ROUGE-L	&\cellcolor{third}48.27&42.58& 44.19	& \cellcolor{second}48.33 & 47.52&47.92&45.12&44.01&45.59&47.50&46.29&\cellcolor{first}\textbf{51.64}\\ 
			&METEOR		&44.63&42.97& 47.55	& \cellcolor{second}49.96 & 47.88&\cellcolor{third}49.36&44.32&45.52&44.80&48.32&45.94&\cellcolor{first}\textbf{52.76}\\ 
			&BERTScore	&65.11&61.19& 63.95	& \cellcolor{second}66.54	& \cellcolor{third}66.01&65.93&62.02&63.34&62.28&65.51&63.14&\cellcolor{first}\textbf{67.65}\\
			\hline
			\multirow{5}{*}{Heart} 
			&BLEU-1		&71.76&57.06& 73.52  & \cellcolor{third}77.45 & 77.43&72.25	&70.04&\cellcolor{second}77.72&72.24&\cellcolor{first}\textbf{77.97}	&72.18&76.48\\
			&ROUGE-1	&74.74&64.65& 74.17  & \cellcolor{third}75.94 & 73.49&75.06	&71.71&75.09&73.81&\cellcolor{second}76.82			&75.94&\cellcolor{first}\textbf{78.18}\\
			&ROUGE-L	&77.12&65.50& 77.04  & \cellcolor{third}79.92 & 78.35&77.56	&74.99&79.54&76.76&\cellcolor{second}80.35			&77.84&\cellcolor{first}\textbf{80.64}\\ 
			&METEOR		&63.13&50.86& 66.32  & \cellcolor{second}73.30 & 72.07&63.68	&61.77&\cellcolor{third}73.02&64.67&\cellcolor{first}\textbf{73.78}	&63.48&70.73\\ 
			&BERTScore	&87.64&82.52& 87.82  & \cellcolor{second}89.94 & 89.40&87.90	&85.27&\cellcolor{third}89.86&86.20&\cellcolor{first}\textbf{90.12}	&87.91&89.80\\
			\hline
			\multirow{5}{*}{Kidney} 
			&BLEU-1		&45.72&39.79& 48.51 & \cellcolor{second}52.09	& 49.54&43.80&41.71&44.84&42.27&\cellcolor{third}49.70&43.79&\cellcolor{first}\textbf{59.23}\\
			&ROUGE-1	&57.69&46.28& 54.77 & \cellcolor{third}58.32	& \cellcolor{second}59.49&49.23&51.86&48.23&52.59&55.75&50.56&\cellcolor{first}\textbf{65.78}\\
			&ROUGE-L	&47.66&38.36& 43.73 & \cellcolor{third}49.26	& \cellcolor{second}50.82&41.95&43.53&40.38&44.25&47.47&42.07&\cellcolor{first}\textbf{57.14}\\ 
			&METEOR		&44.15&39.68& 47.70 & \cellcolor{second}52.57	& \cellcolor{third}51.57&42.82&42.81&43.03&43.38&49.87&43.17&\cellcolor{first}\textbf{59.78}\\
			&BERTScore	&67.96&58.86& 66.86	& \cellcolor{second}70.63	& \cellcolor{third}70.23&62.65&63.37&63.08&63.91&68.31&63.20&\cellcolor{first}\textbf{75.89}\\
			\hline
			\multirow{5}{*}{Liver} 
			&BLEU-1		&47.00&40.19	& 49.40 & 53.11	& \cellcolor{second}53.87 &	40.48&43.10&44.28&43.68&\cellcolor{third}53.64&50.91&\cellcolor{first}\textbf{58.01}\\
			&ROUGE-1	&\cellcolor{third}61.90&44.34	& 55.02	& 60.95	& 60.59 &	51.93&53.77&47.41&54.53&59.82&\cellcolor{second}63.57&\cellcolor{first}\textbf{67.06}\\
			&ROUGE-L	&\cellcolor{third}58.28&40.18	& 50.85	& 56.67	& 56.35 &	46.78&50.89&43.39&51.68&53.89&\cellcolor{second}59.30&\cellcolor{first}\textbf{62.40}\\ 
			&METEOR		&48.93&38.85	& 47.82	& 52.87	& \cellcolor{third}53.43 &	41.54&45.63&41.52&46.39&53.01&\cellcolor{second}53.59&\cellcolor{first}\textbf{58.44}\\
			&BERTScore	&70.86&58.82	& 67.73	& 71.58	& \cellcolor{third}72.22 &	65.98&64.66&62.33&65.22&70.91&\cellcolor{second}72.51&\cellcolor{first}\textbf{75.68}\\
			\hline
			\multirow{5}{*}{Vessel} 
			&BLEU-1		&39.10			&29.75	& \cellcolor{third}41.15	& 40.65& \cellcolor{first}\textbf{42.54}	&37.62	&16.12&37.39&29.56&\cellcolor{second}42.53		 &31.64&29.54\\
			&ROUGE-1	&\cellcolor{first}\textbf{65.14} &45.68	& 53.40	& \cellcolor{second}63.78& 60.81			&60.63	&20.06&55.57&45.65&\cellcolor{third}63.51		 &42.14&38.15\\
			&ROUGE-L	&\cellcolor{first}\textbf{58.36}	&39.15	& 46.56	& \cellcolor{second}56.98& 53.79 			&55.38	&13.36&48.43&39.06&\cellcolor{third}56.86		 &35.95&32.29\\ 
			&METEOR		&\cellcolor{second}50.80			&36.98	& 47.39	& \cellcolor{second}50.80& \cellcolor{third}50.77			&46.50	&15.27&45.06&36.97&\cellcolor{first}\textbf{51.28}&34.41&30.38\\
			&BERTScore	&\cellcolor{first}\textbf{74.00}	&54.97	& 62.27	& \cellcolor{second}72.88& 69.03 			&70.01	&28.23&64.32&55.08&\cellcolor{third}71.74		 &51.76&48.07\\
			\hline
			\multirow{5}{*}{Thyroid} 
			&BLEU-1		&23.17&26.79	& 39.68	& \cellcolor{second}45.59	& 41.50 &	38.25&44.27&44.23&44.00&39.37&\cellcolor{third}45.13&\cellcolor{first}\textbf{50.55}\\
			&ROUGE-1	&38.35&35.21	& 48.72	& \cellcolor{second}54.60	& 47.51 &	51.03&\cellcolor{third}53.30&52.93&52.92&48.91&52.65&\cellcolor{first}\textbf{59.10}\\
			&ROUGE-L	&30.50&27.81	& 40.54	& \cellcolor{second}45.50	& 40.30 &	41.00&\cellcolor{third}44.45&44.10&43.99&39.68&\cellcolor{third}44.47&\cellcolor{first}\textbf{51.16}\\ 
			&METEOR		&25.52&26.57	& 40.99	& 44.85	& 39.41 &	37.63&\cellcolor{second}45.51&43.95&\cellcolor{third}45.13&38.13&44.21&\cellcolor{first}\textbf{49.87}\\
			&BERTScore	&50.13&49.51	& 59.75	& \cellcolor{third}66.06	& 61.15 &	\cellcolor{second}66.59&63.86&64.89&63.54&61.36&64.11&\cellcolor{first}\textbf{69.54}\\
			\hline
			\multirow{5}{*}{Average} 
			&BLEU-1		&42.14	& 40.51	& 50.16	& \cellcolor{second}52.75 &\cellcolor{third}51.76	&46.29&44.15&49.47&46.32&51.56&48.57&\cellcolor{first}\textbf{53.87}\\
			&ROUGE-1	&56.38	& 48.86	& 57.18	& \cellcolor{second}61.59 &59.76	&58.24&51.83&56.17&55.80&\cellcolor{third}60.14&57.18&\cellcolor{first}\textbf{61.69}\\
			&ROUGE-L	&50.61	& 43.27	& 50.65	& \cellcolor{second}55.55 &\cellcolor{third}54.09	&52.01&46.34&50.32&50.27&53.62&51.32&\cellcolor{first}\textbf{55.78}\\ 
			&METEOR		&43.37	& 40.17	& 49.65	& \cellcolor{first}\textbf{53.38} &\cellcolor{third}52.14	&46.98&43.39&48.80&46.92&51.69&47.80&\cellcolor{second}53.16\\
			&BERTScore	&66.72	& 61.83	& 68.39	& \cellcolor{first}\textbf{72.57} &\cellcolor{third}71.27	&70.18&62.42&68.48&66.36&71.19&67.64&\cellcolor{second}71.38\\
			\hline
		\end{tabular}
	}
	\caption{Comparison results for report generation.}
	\label{table2}
\end{table*}
\subsection{Implementation Details}
Following LLaVA-Med’s protocol, the two-stage training framework adopts phase-specific datasets. Stage I initializes newly introduced MoE modules using 208,941 clinical reports and 1.47 million ultrasound key frames, exclusively training them to capture domain-specific visual-textual patterns without modifying the base model’s pre-trained parameters. Stage II performs Cooperative Instruction Fine-tuning on 1.8 million instruction-following samples, integrating the base model and MoE modules to enhance instruction understanding and response capabilities. For reproducibility, the model is evaluated on a held-out test set (27,577 ultrasound images and 3,000 reports), with greedy decoding during inference ensuring deterministic and reproducible outputs. More details are provided in the Appendix~\ref{sec:appendix_b}.
\subsection{Results}
\begin{table*}[!ht]
	\small 
	\centering
	\resizebox{\linewidth}{!}{%
		\begin{tabular}{c|c|cccccccccccc}
			\hline
			Anatomical	&Metric$\uparrow$&\makecell{Medgemma\\(4B)}&\makecell{LLaVA1.5\\(7B)}&\makecell{HuatuoGPT-Vision\\(7B)}& \makecell{Lingshu\\(7B)} & \makecell{LLaVA-OneVision-\\(7B)}&\makecell{Qwen2-VL-\\(7B)}&\makecell{LLaVA-Med (7B)}&\makecell{Qwen2.5-VL\\(7B)}&\makecell{LLaVa-NeXT\\(13B)}&\makecell{Gemma-3\\(12B)}&\makecell{EchoVLM\\(3B)} &EchoVLM(11B) \\  
			\hline
			\multirow{5}{*}{Breast} 
			&BLEU-1		&61.94&54.79& 65.29	& \cellcolor{second}70.37	& 65.46 &67.63&\cellcolor{third}69.84&68.09&68.82&67.91&67.92		 &\cellcolor{first}\textbf{71.36}\\ 
			&ROUGE-1 	&77.42&70.42& 77.00	& \cellcolor{second}78.74	& 75.78	&78.13&\cellcolor{third}78.37&77.61&77.61&75.19&78.11		 &\cellcolor{first}\textbf{80.77}\\
			&ROUGE-L  	&70.17&58.71& 69.93	& \cellcolor{second}72.98	& 67.22	&71.38&\cellcolor{third}72.30&69.46&71.21&68.28&70.00		 &\cellcolor{first}\textbf{76.04}\\ 
			&METEOR  	&65.99&64.29& 71.25	& \cellcolor{second}75.30	& 73.99	&72.88&74.34&74.92&73.53&72.79&\cellcolor{first}\textbf{75.75}&\cellcolor{third}74.99\\
			&BERTScore	&77.81&69.20& 77.07	& \cellcolor{second}79.23	& 75.70	&78.32&\cellcolor{third}78.72&77.34&77.98&76.70&78.03		 &\cellcolor{first}\textbf{81.68}\\
			\hline
			\multirow{5}{*}{Gynecology} 
			&BLEU-1		&28.38&29.77& 39.59 &36.46	& 33.46 &	\cellcolor{third}40.46&37.80&36.37&38.21&37.42&\cellcolor{second}43.32&\cellcolor{first}\textbf{48.15}\\
			&ROUGE-1  	&47.43&38.66& 48.71 &45.70	& 45.33 &	53.13&53.56&46.39&\cellcolor{third}53.83&48.87&\cellcolor{second}56.24&\cellcolor{first}\textbf{62.82}\\
			&ROUGE-L  	&42.44&36.29& 43.74 &42.01	& 40.79	&	48.44&49.44&42.25&\cellcolor{third}49.74&44.99&\cellcolor{second}52.15&\cellcolor{first}\textbf{58.25}\\ 
			&METEOR  	&33.43&36.92& 44.62 &42.28	& 40.16	&	\cellcolor{third}47.32&45.82&43.53&46.19&44.12&\cellcolor{second}50.86&\cellcolor{first}\textbf{55.95}\\
			&BERTScore	&52.65&43.56& 53.50 &50.97	& 50.18	&	57.81&58.53&51.38&\cellcolor{third}58.82&53.99&\cellcolor{second}60.03&\cellcolor{first}\textbf{66.58}\\
			\hline
			\multirow{5}{*}{Heart} 
			&BLEU-1		&55.99&42.41& 55.32& 64.60	& 61.00 &	66.26&\cellcolor{third}67.77&60.12&\cellcolor{second}69.31&65.04&66.30&\cellcolor{first}\textbf{69.62}\\
			&ROUGE-1  	&71.58&50.32& 62.77& 76.49	& 70.95	&	77.16&\cellcolor{third}79.13&71.90&\cellcolor{second}79.94&75.89&78.74&\cellcolor{first}\textbf{81.33}\\
			&ROUGE-L  	&64.63&44.81& 55.69& 69.31	& 63.39	&	69.78&71.26&64.78&\cellcolor{second}72.55&68.57&\cellcolor{third}71.47&\cellcolor{first}\textbf{74.35}\\ 
			&METEOR  	&59.00&43.54& 59.43& 67.68	& 64.16	&	68.59&\cellcolor{third}70.26&63.46&\cellcolor{second}71.63&68.02&68.91&\cellcolor{first}\textbf{72.63}\\ 
			&BERTScore	&73.15&55.58& 67.65& 76.34	& 71.59	&	77.63&\cellcolor{third}78.86&72.96&\cellcolor{second}79.88&76.02&78.06&\cellcolor{first}\textbf{79.94}\\
			\hline
			\multirow{5}{*}{Kidney} 
			&BLEU-1		&66.68&44.09& 62.52 & 69.15	& \cellcolor{third}71.76 &68.71&70.76&63.05&70.74&69.95&\cellcolor{second}74.12&\cellcolor{first}\textbf{77.56}\\
			&ROUGE-1  	&77.06&52.81& 70.77 & 76.37	& \cellcolor{third}79.32	&76.87&78.10&71.77&78.00&78.01&\cellcolor{second}81.70&\cellcolor{first}\textbf{83.42}\\
			&ROUGE-L  	&67.55&44.51& 59.51 & 67.23	& \cellcolor{third}70.77	&67.23&68.58&61.64&68.59&68.02&\cellcolor{second}72.89&\cellcolor{first}\textbf{74.86}\\
			&METEOR  	&67.15&52.09& 68.75 & 73.26	& \cellcolor{third}75.19	&72.57&74.80&69.18&74.75&74.75&\cellcolor{second}76.56&\cellcolor{first}\textbf{80.90}\\
			&BERTScore	&81.51&60.40& 76.07 & 80.80	& \cellcolor{third}83.46	&81.46&82.17&77.08&82.13&81.78&\cellcolor{second}84.71&\cellcolor{first}\textbf{87.03}\\
			\hline
			\multirow{5}{*}{Liver} 
			&BLEU-1		&60.39&50.19& 61.98 & \cellcolor{third}67.11	& 66.60 &	63.23&62.42&60.77&62.93&58.88&\cellcolor{second}68.45&\cellcolor{first}\textbf{74.23} \\
			&ROUGE-1 	&70.87&59.10& 69.42 & \cellcolor{third}73.84	& 73.02 &	73.63&70.45&68.58&70.80&65.89&\cellcolor{second}75.85&\cellcolor{first}\textbf{79.87}  \\
			&ROUGE-L  	&64.49&52.72& 63.56 & \cellcolor{third}69.00	& 68.26 &	66.49&64.65&62.91&65.14&58.93&\cellcolor{second}69.32&\cellcolor{first}\textbf{75.06}\\ 
			&METEOR  	&63.96&58.21& 69.15 & \cellcolor{third}72.34	& 71.22 &	68.75&71.10&70.02&71.48&68.71&\cellcolor{second}73.77&\cellcolor{first}\textbf{78.49}\\
			&BERTScore	&73.55&63.36& 73.24 & \cellcolor{third}77.40	& 77.01 &	76.14&73.65&73.06&73.96&70.62&\cellcolor{second}77.81&\cellcolor{first}\textbf{82.85}\\
			\hline
			\multirow{5}{*}{Vessel} 
			&BLEU-1		&\cellcolor{second}53.41&31.68& 49.23	& 47.86	& \cellcolor{third}49.65 &	\cellcolor{first}\textbf{59.56}&27.54&42.85&35.88&46.86&38.70&36.27\\
			&ROUGE-1 	&\cellcolor{second}66.86&47.47& 57.85	& \cellcolor{third}60.58	& 60.12 &	\cellcolor{first}\textbf{70.69}&42.75&56.73&51.35&58.23&51.00&47.63\\
			&ROUGE-L  	&\cellcolor{second}65.32&44.99& 55.49	& \cellcolor{third}59.09	& 58.31 &	\cellcolor{first}\textbf{69.54}&36.99&54.51&49.05&56.41&46.33&44.08\\
			&METEOR  	&\cellcolor{second}59.53&40.97& \cellcolor{third}56.04	& 53.12	& 55.46 &	\cellcolor{first}\textbf{64.98}&45.73&53.06&44.78&51.32&45.06&40.74\\
			&BERTScore	&\cellcolor{second}74.45&58.16& 66.90	& \cellcolor{third}69.33	& 67.44 &	\cellcolor{first}\textbf{76.82}&46.97&65.74&61.19&67.10&58.43&53.86\\
			\hline
			\multirow{5}{*}{Thyroid} 
			&BLEU-1		&49.51&32.60& 54.82	& 52.93	& 53.03 &	51.86&52.27&\cellcolor{third}55.24&52.17&52.38&\cellcolor{second}57.23&\cellcolor{first}\textbf{62.49}\\
			&ROUGE-1 	&59.95&42.84& \cellcolor{third}65.57	& 63.85	& 62.01 &	65.01&61.33&63.32&60.89&59.82&\cellcolor{second}66.30&\cellcolor{first}\textbf{71.74}\\
			&ROUGE-L  	&54.39&36.84& \cellcolor{second}61.37	& 58.85	& 56.08 &	59.56&57.12&58.82&56.78&53.76&\cellcolor{third}60.97&\cellcolor{first}\textbf{67.46}\\ 
			&METEOR  	&55.29&44.49& \cellcolor{third}63.79	& 62.23	& 53.03 &	61.06&61.90&63.65&61.47&61.17&\cellcolor{second}64.29&\cellcolor{first}\textbf{70.03}\\
			&BERTScore	&65.05&48.68& \cellcolor{second}70.74	& 68.90	& 66.89 &	69.69&67.18&68.36&66.81&65.12&\cellcolor{third}70.49&\cellcolor{first}\textbf{76.11}\\
			\hline
			\multirow{5}{*}{Average} 
			&BLEU-1		&53.76&40.79&55.54& 58.35&57.28&\cellcolor{second}59.67&55.49&55.21&56.87&56.92&\cellcolor{third}59.43&\cellcolor{first}\textbf{62.81}\\
			&ROUGE-1 	&67.31&51.66&64.58& 67.94&66.65&\cellcolor{second}70.66&66.24&65.19&67.49&65.99&\cellcolor{third}69.71&\cellcolor{first}\textbf{72.51}\\
			&ROUGE-L  	&61.28&45.55&58.47& 62.64&60.69&\cellcolor{second}64.63&60.05&59.20&61.87&59.85&\cellcolor{third}63.30&\cellcolor{first}\textbf{67.16}\\ 
			&METEOR  	&57.76&48.64&61.86& 63.74&61.89&\cellcolor{second}65.16&63.42&62.55&63.40&62.98&\cellcolor{third}65.03&\cellcolor{first}\textbf{67.68}\\
			&BERTScore	&71.17&56.99&69.31& 71.85&70.32&\cellcolor{second}73.98&69.44&69.42&71.54&70.19&\cellcolor{third}72.51&\cellcolor{first}\textbf{75.44}\\
			\hline
		\end{tabular}
	}
	\caption{Comparison results for ultrasound diagnosis.}
	\label{table3}
\end{table*}
\begin{table*}[!ht]
	\small 
	\centering
	\resizebox{\linewidth}{!}{%
		\begin{tabular}{c|c|cccccccccccc}
			\hline
			Anatomical	&Metric$\uparrow$&\makecell{Medgemma\\(4B)}&\makecell{LLaVA1.5\\(7B)}&\makecell{HuatuoGPT-Vision\\(7B)}& \makecell{Lingshu\\(7B)} & \makecell{LLaVA-OneVision-\\(7B)}&\makecell{Qwen2-VL-\\(7B)}&\makecell{LLaVA-Med (7B)}&\makecell{Qwen2.5-VL\\(7B)}&\makecell{LLaVa-NeXT\\(13B)}&\makecell{Gemma-3\\(12B)}&\makecell{EchoVLM\\(3B)} &EchoVLM(11B) \\  
			\hline
			\multirow{5}{*}{Breast} 
			&BLEU-1		&25.03&25.81& 32.35	& \cellcolor{second}36.43	& \cellcolor{first}\textbf{36.79} & 33.60&25.61&\cellcolor{third}36.62		 &28.31&35.86&34.09&35.80\\ 
			&ROUGE-1 	&35.36&38.82& 38.25	& \cellcolor{second}44.26	& 43.31			 & 42.61&36.54&\cellcolor{first}\textbf{44.75}&36.97&43.16&43.82&\cellcolor{third}44.55\\
			&ROUGE-L  	&27.18&30.72& 29.82	& \cellcolor{second}36.13	& 35.03			 & 34.73&26.45&\cellcolor{first}\textbf{36.75}&28.48&34.57&35.62&\cellcolor{third}36.31\\ 
			&METEOR  	&23.73&23.49& 29.37	& \cellcolor{second}30.77	& \cellcolor{first}\textbf{31.47} & 28.19&22.00&\cellcolor{third}31.07		 &25.59&29.26&29.09&29.76\\
			&BERTScore	&42.22&46.86& 46.26	& \cellcolor{second}54.37	& 54.33 		 & 52.73&47.50&\cellcolor{first}\textbf{55.25}&44.34&53.45&53.67&\cellcolor{third}54.46\\
			\hline
			\multirow{5}{*}{Gynecology} 
			&BLEU-1		&23.33&21.58 &30.17	& \cellcolor{second}32.76	& 29.88 &28.02&24.46&31.64&26.23&\cellcolor{first}\textbf{33.12}&25.74&\cellcolor{third}25.75\\
			&ROUGE-1  	&38.64&32.93 &38.24	& \cellcolor{third}41.22	& 39.01 &40.40&38.03&40.12&39.95&\cellcolor{second}41.25&39.89&\cellcolor{first}\textbf{42.10}\\
			&ROUGE-L  	&30.01&24.20 &28.62	& \cellcolor{third}31.64	& 30.01	&31.69&28.87&31.24&31.50&\cellcolor{second}31.61&31.09&\cellcolor{first}\textbf{33.30}\\ 
			&METEOR  	&22.52&20.29 &25.99	&\cellcolor{second}29.20	& 27.13	&25.21&22.76&28.07&24.01&\cellcolor{first}\textbf{30.04}&24.2&\cellcolor{third}25.59\\
			&BERTScore	&48.24&41.12 &47.73	& \cellcolor{third}50.86	& 48.81	&49.84&49.35&49.77&49.49&\cellcolor{second}51.27&49.13&\cellcolor{first}\textbf{52.21}\\
			\hline
			\multirow{5}{*}{Heart} 
			&BLEU-1		&21.46&21.58&27.77& \cellcolor{third}27.98		  & \cellcolor{first}\textbf{30.43} 	&22.96&21.20&28.18&23.75&\cellcolor{second}28.88			&20.10& 23.55\\
			&ROUGE-1  	&34.39&30.23&35.14& \cellcolor{first}\textbf{37.00}& \cellcolor{third}35.50			&34.92&34.04&36.43&35.71&\cellcolor{second}36.95			&35.60&35.93\\
			&ROUGE-L  	&25.88&22.56&26.54& \cellcolor{first}\textbf{28.87}& \cellcolor{third}27.69 			&26.75&24.92&28.51&27.53&\cellcolor{second}28.84			&27.28& 27.62\\
			&METEOR  	&21.84&20.66&24.85& \cellcolor{third}25.72		  & \cellcolor{first}\textbf{27.76} 	&22.87&20.22&25.73&23.06&\cellcolor{second}26.43			&22.10& 23.86\\ 
			&BERTScore	&43.79&38.73&45.21& \cellcolor{second}46.92		  & 46.25			&45.37&44.78&46.07&45.50&\cellcolor{first}\textbf{47.23}	&45.47&\cellcolor{third}46.38\\
			\hline
			\multirow{5}{*}{Kidney} 
			&BLEU-1		&25.39&24.27&29.62& \cellcolor{first}\textbf{33.49}	& \cellcolor{third}31.30	&28.99&24.16&32.08&27.01&\cellcolor{second}33.10			&26.85&31.59\\
			&ROUGE-1  	&39.60&35.74&39.16& \cellcolor{first}\textbf{42.75}	& \cellcolor{third}41.46	&41.23&37.07&41.51&40.45&\cellcolor{second}42.57			&40.98&42.47\\
			&ROUGE-L  	&30.69&27.00&29.89& \cellcolor{first}\textbf{33.61}	& \cellcolor{third}32.46	&32.25&27.14&32.46&31.27&\cellcolor{second}33.31			&31.76&33.59\\
			&METEOR  	&24.16&23.23&25.91& \cellcolor{second}30.30			& 28.58	&26.65&22.39&29.38&25.30&\cellcolor{first}\textbf{30.56}	&25.30&\cellcolor{third}28.48\\
			&BERTScore	&48.79&44.78&49.09& \cellcolor{first}\textbf{52.93}	& 51.67	&51.10&48.49&51.67&49.86&\cellcolor{first}52.93	&49.91&\cellcolor{second}52.24\\
			\hline
			\multirow{5}{*}{Liver} 
			&BLEU-1		&29.24&24.82&	35.95	& \cellcolor{second}35.70	& 35.30 		 &	32.00&27.84&34.21&30.45&33.18&31.43&\cellcolor{first}\textbf{37.26}\\
			&ROUGE-1 	&39.75&34.46&	42.16	& \cellcolor{second}42.51	& 40.65 		 &	41.24&37.73&40.86&40.13&39.55&41.59&\cellcolor{first}\textbf{44.36}\\
			&ROUGE-L  	&29.72&25.39&	32.03	& \cellcolor{second}32.66	& 30.66 		 &	31.67&27.20&31.44&30.26&29.09&31.96&\cellcolor{first}\textbf{34.37}\\
			&METEOR  	&25.61&22.26&	31.26	& \cellcolor{third}30.50	& 30.96			 &	27.16&23.34&29.10&25.94&27.79&27.05&\cellcolor{first}\textbf{31.17}\\
			&BERTScore	&49.05&43.10&	52.44	& \cellcolor{second}52.59	& 51.34 		 &	51.19&48.94&51.13&49.45&49.39&51.02&\cellcolor{first}\textbf{54.20}\\
			\hline
			\multirow{5}{*}{Vessel} 
			&BLEU-1		&28.83&20.97&32.10& \cellcolor{first}\textbf{35.24}	& \cellcolor{second}34.27 &	29.71&24.30&32.99&21.48&\cellcolor{third}34.25&26.46&27.06\\
			&ROUGE-1 	&40.24&33.51&38.74& \cellcolor{first}\textbf{42.17}	& 39.17 &	40.70&35.38&40.56&33.85&\cellcolor{second}41.32&36.75&\cellcolor{third}37.82\\
			&ROUGE-L  	&30.82&24.89&29.63& \cellcolor{first}\textbf{33.56}	& 31.30 &	32.20&25.30&31.87&25.30&\cellcolor{second}32.61&27.72&\cellcolor{third}28.79\\
			&METEOR  	&25.85&20.77&26.52& \cellcolor{first}\textbf{29.40}	& \cellcolor{second}28.94 &	25.63&22.02&28.65&20.94&\cellcolor{third}29.14&24.18&24.55\\
			&BERTScore	&51.25&44.56&50.54& \cellcolor{first}\textbf{53.65}	& 51.50 &	51.86&47.33&51.92&44.88&\cellcolor{second}53.06&47.87&\cellcolor{third}49.42\\
			\hline
			\multirow{5}{*}{Thyroid} 
			&BLEU-1		&25.14&26.30&	31.93	& \cellcolor{third}33.41	& \cellcolor{first}\textbf{34.77} &	30.69	&25.07&\cellcolor{second}33.65&25.81&31.73&28.3&29.92\\
			&ROUGE-1 	&35.98&36.54&	38.84	& \cellcolor{second}42.17	& 41.67 		 &	41.78	&36.20&41.72&35.43&40.39&40.97&\cellcolor{first}\textbf{42.31}\\
			&ROUGE-L  	&28.36&29.17&	30.99	& \cellcolor{second}34.44	& 34.23			 &	34.25	&27.35&33.89&27.61&32.45&33.34&\cellcolor{first}\textbf{34.74}\\ 
			&METEOR  	&24.31&23.74&	28.95	& \cellcolor{third}29.74	& \cellcolor{first}\textbf{30.76} &	27.38	&22.28&\cellcolor{second}29.85&24.48&27.39&26.43&27.86\\
			&BERTScore	&44.54&47.30&	48.35	& \cellcolor{second}53.29	& \cellcolor{first}\textbf{53.74} &	52.73	&47.90&53.07&44.14&51.50&51.80&\cellcolor{third}53.38\\
			\hline
			\multirow{5}{*}{Average} 
			&BLEU-1		&25.49&  23.62	& 31.41	&\cellcolor{first}33.57&	\cellcolor{second}33.25&29.42	&24.66&32.77&26.15&\cellcolor{third}32.87&27.57&30.13\\
			&ROUGE-1 	&37.71&	34.60	& 38.65	&\cellcolor{second}41.73&	40.11&40.41	&36.43&40.85&37.50&\cellcolor{third}40.74&39.94&\cellcolor{first}\textbf{41.36}\\
			&ROUGE-L  	&28.95& 26.28	& 29.65	&\cellcolor{second}32.99&	31.63&31.93	&26.75&32.31&28.85&\cellcolor{third}31.78&31.25&\cellcolor{first}\textbf{32.67}\\ 
			&METEOR  	&24.00&	22.06	& 27.55	&\cellcolor{first}29.38&	\cellcolor{second}29.37&26.16	&22.14&28.84&24.19&\cellcolor{third}28.66&25.48&27.32\\
			&BERTScore	&46.84&	43.78	& 48.52	&\cellcolor{first}52.09&	51.09&50.69	&47.76&51.27&46.81&\cellcolor{third}51.26&49.84&\cellcolor{second}51.76\\
			\hline
		\end{tabular}
	}
	\caption{Comparison results for VQA.}
	\label{table4}
\end{table*}
\subsubsection{Report Generation} Table \ref{table2} compares 12 VLMs (general and medical-specific) across seven anatomical systems using five key metrics. EchoVLM (11B) is the core highlight, consistently outperforming counterparts (e.g., Qwen2-VL~\cite{Qwen2-VL}, Qwen2.5-VL~\cite{Qwen2.5-VL}, Gemma-3~\cite{Gemma-3}) and achieving SOTA in multiple scenarios.
It tops BERTScore in four systems (breast:73.00, kidney:75.89, etc.), leads ROUGE-1/ROUGE-L in breast, kidney, liver, and thyroid, and ranks first in BLEU-1 and METEOR for most domains. Its average scores are outstanding, with the highest scores in BLEU-1 (53.87), ROUGE-1 (61.69), and ROUGE-L (55.78), and the second-highest in METEOR and BERTScore. This proves its strong domain-specific performance and generalizability.
Even the smaller EchoVLM (3B) performs admirably (e.g., 3rd in breast BLEU-1/ROUGE-1). In contrast, compared with other models adopting full-parameter fine-tuning, this outstanding performance of EchoVLM demonstrates the high efficiency of leveraging MoE to inject domain knowledge.

Traditional n-gram overlap and text similarity metrics (BLEU, ROUGE, BERTScore) are widely used for report generation evaluation but correlate imperfectly with clinical correctness, as they favor surface-level text matching over accurate medical concepts, attributes and critical findings. Therefore, we supplement our experiments with fine-grained entity-level metrics, extracting anatomical structures, findings, attributes and statuses via a clinically validated extractor and computing four standard classification metrics. Complete results are provided in Appendix~\ref{sec:supp_entity_eval}.

\subsubsection{Ultrasound Diagnosis}
Ultrasound diagnosis is a highly specialized synthesis of sonographic findings, requiring precise integration of anatomical and imaging findings with high professional accuracy. To validate EchoVLM’s efficacy in this task, we evaluated it against other models across seven anatomical systems. Table \ref{table3} shows EchoVLM (11B) outperforms other general and medical-specific models in most categories, achieving SOTA in key systems like kidney (BLEU-1: 77.56, BERTScore: 87.03) and liver (ROUGE-1: 79.87, BERTScore: 82.85). It also leads all average metrics (BERTScore:75.44, ROUGE-1:72.51), reflecting strong alignment with clinical terminology and superior diagnostic reasoning.
\subsubsection{Vision Question Answering}
Ultrasound VQA, as an open-ended question task, typically demands models with hierarchical comprehension capabilities to satisfy heterogeneous clinical requirements. Such multi-level proficiency is imperative for decision support for both radiologists and referring physicians. Hence, comparative experiments were conducted across seven anatomical systems and five metrics, with results reported in Table \ref{table4}. Collectively, EchoVLM exhibits balanced performance: it surpasses most general and medical-specific models in 21 of 35 metric-anatomy combinations and ranks among the top 3 in an additional 8. Notably, these results demonstrate the effectiveness of leveraging a MoE architecture in achieving effective knowledge transfer to domain-specific ultrasound VQA tasks. Although our model lags behind Lingshu in certain metrics, its overall performance is superior to Lingshu, with more details provided in the Appendix~\ref{sec:appendix_b}.
\subsection{Ablation Study}
\subsubsection{Impact of the share expert}
The ablation study in Table~\ref{ablation_share_expert} demonstrates that integrating shared experts into the MoE framework improves performance for medical report generation and ultrasound diagnosis, with BLEU-1 gains of +4.58 and +3.48, ROUGE-L increases of +3.45 and +5.49, and BERTScore improvements of +1.27 and +4.06, respectively. These results indicate that shared experts enhance cross-task knowledge transfer and multimodal coherence in tasks requiring complex semantic synthesis.
\begin{table}[ht]
	\small 
	\centering
	\begin{tabular}{c|c|cc}
		\hline
		Task	&Metric$\uparrow$&\makecell{w/o \\Share\\ Expert}&\makecell{w \\Share\\ Expert}\\ 
		\hline
		\multirow{5}{*}{\makecell{Report \\Generation} }
		&BLEU-1		& 49.29 & 53.87 \textbf{\textcolor{lightred}{(+4.58)}} \\ 
		&ROUGE-1 	& 58.42	& 61.69 \textbf{\textcolor{lightred}{(+3.27)}}\\
		&ROUGE-L  	& 52.33	& 55.78 \textbf{\textcolor{lightred}{(+3.45)}}\\ 
		&METEOR  	& 49.54	& 53.16 \textbf{\textcolor{lightred}{(+3.62)}} \\
		&BERTScore	& 70.11	& 71.38 \textbf{\textcolor{lightred}{(+1.27)}}\\
		\hline
		\multirow{5}{*}{\makecell{Ultrasound\\ Diagnosis}} 
		&BLEU-1		& 59.33 & 62.81 \textbf{\textcolor{lightred}{(+3.48)}} \\
		&ROUGE-1  	& 67.46 & 72.51 \textbf{\textcolor{lightred}{(+5.05)}} \\
		&ROUGE-L  	& 61.67 & 67.16 \textbf{\textcolor{lightred}{(+5.49)}}\\ 
		&METEOR  	& 64.92 & 67.68 \textbf{\textcolor{lightred}{(+2.76)}}\\
		&BERTScore	& 71.38 & 75.44  \textbf{\textcolor{lightred}{(+4.06)}}	\\
		\hline
		\multirow{5}{*}{VQA} 
		&BLEU-1		& 33.15 &30.13 \textbf{\textcolor{lightgreen}{(-3.02)}}\\
		&ROUGE-1  	& 38.61 &41.36 \textbf{\textcolor{lightred}{(+2.75)}}\\
		&ROUGE-L  	& 30.49 &32.67 \textbf{\textcolor{lightred}{(+2.18)}}\\
		&METEOR  	& 29.09 &27.32 \textbf{\textcolor{lightgreen}{(-1.77)}}\\ 
		&BERTScore	& 49.60 &51.76 \textbf{\textcolor{lightred}{(+2.16)}}\\
		\hline
	\end{tabular}
	\caption{Ablation study of the shared expert.}
	\label{ablation_share_expert}
\end{table}
\subsubsection{Impact of the Top-K Routing.}
To investigate the impact of expert activation mechanisms within the MoE architecture, we implement two widely adopted routing strategies: Top-1 and Top-2 activation. As shown in Table~\ref{ablation_topk}, ablation studies demonstrate that Top-2 routing generally achieves superior performance in complex multimodal tasks compared to Top-1 routing, although the magnitude of improvement varies across task characteristics. For report generation tasks, Top-2 routing enhances all evaluation metrics (e.g., +3.94 in BLEU-1, +2.81 in ROUGE-L, and +3.29 in METEOR). For ultrasound diagnostic tasks, it delivers remarkable gains across key metrics, including +4.10 in BLEU-1, +4.76 in ROUGE-1, and a substantial +5.00 in ROUGE-L. These results indicate that activating two specialized experts facilitates more effective knowledge integration and contextual understanding.
\begin{table}[ht]
	\small 
	\centering
	\begin{tabular}{c|c|cc}
		\hline
		Task(4 Experts)&Metric$\uparrow$&\makecell{Top1}&\makecell{Top2}\\ 
		\hline
		\multirow{5}{*}{\makecell{Report \\Generation} }
		&BLEU-1		& 49.93 & 53.87 \textbf{\textcolor{lightred}{(+3.94)}} \\ 
		&ROUGE-1 	& 58.80	& 61.69 \textbf{\textcolor{lightred}{(+2.89)}}\\
		&ROUGE-L  	& 52.97	& 55.78 \textbf{\textcolor{lightred}{(+2.81)}}\\ 
		&METEOR  	& 49.87	& 53.16 \textbf{\textcolor{lightred}{(+3.29)}} \\
		&BERTScore	& 70.54	& 71.38 \textbf{\textcolor{lightred}{(+0.84)}}\\
		\hline
		\multirow{5}{*}{\makecell{Ultrasound\\ Diagnosis}} 
		&BLEU-1		& 58.71 & 62.81 \textbf{\textcolor{lightred}{(+4.10)}} \\
		&ROUGE-1  	& 67.75 & 72.51 \textbf{\textcolor{lightred}{(+4.76)}} \\
		&ROUGE-L  	& 62.16 & 67.16 \textbf{\textcolor{lightred}{(+5.00)}}\\ 
		&METEOR  	& 64.73 & 67.68 \textbf{\textcolor{lightred}{(+2.94)}}\\
		&BERTScore	& 72.20 & 75.44  \textbf{\textcolor{lightred}{(+3.23)}}	\\
		\hline
		\multirow{5}{*}{VQA} 
		&BLEU-1		& 33.58 &30.13 \textbf{\textcolor{lightgreen}{(-3.45)}}\\
		&ROUGE-1  	& 38.81 &41.36 \textbf{\textcolor{lightred}{(+2.55)}}\\
		&ROUGE-L  	& 30.68 & 32.67 \textbf{\textcolor{lightred}{(+1.99)}}\\
		&METEOR  	& 29.60 &27.32 \textbf{\textcolor{lightgreen}{(-2.28)}}\\ 
		&BERTScore	& 49.96 &51.76 \textbf{\textcolor{lightred}{(+1.80)}}\\
		\hline
	\end{tabular}
	\caption{Ablation study of Top-K routing strategies.}
	\label{ablation_topk}
\end{table}
\subsubsection{Impact of the Number of Routing Experts.}
\begin{table}[ht]
	\small 
	\centering
	\resizebox{\linewidth}{!}{%
		\begin{tabular}{c|c|cccc}
			\hline
			Task&Metric$\uparrow$&\makecell{0 Expert}&\makecell{2 Experts}&\makecell{4 Experts}&\makecell{6 Experts}\\ 
			\hline
			\multirow{5}{*}{\makecell{Report \\Generation} }
			&BLEU-1		& 43.72&	50.33	&53.87& 54.03\\
			&ROUGE-1 	& 56.92&	58.75	&61.69& 61.91\\
			&ROUGE-L  	& 49.77&	52.74	&55.78& 56.15\\
			&METEOR  	& 44.76&	50.33	&53.16& 53.59\\
			&BERTScore	& 70.17&	70.51	&71.38& 71.71\\
			\hline
			\multirow{5}{*}{\makecell{Ultrasound\\ Diagnosis}} 
			&BLEU-1		& 58.70 &	59.66	&62.81& 63.23\\
			&ROUGE-1  	& 70.52 &	68.02	&72.51& 73.11\\
			&ROUGE-L  	& 64.52 &	62.45	&67.16& 67.94\\
			&METEOR  	& 64.05 &	65.59	&67.68& 68.47\\
			&BERTScore	& 73.99 &	72.16	&75.44& 75.49\\
			\hline
			\multirow{5}{*}{VQA} 
			&BLEU-1		& 27.64 &	33.52	&30.13&33.73\\
			&ROUGE-1  	& 37.95 &	38.88	&41.36&41.86 \\
			&ROUGE-L  	& 28.96 &	30.69	&32.67&32.68\\
			&METEOR  	& 26.53 &	29.46	&27.32&29.64\\ 
			&BERTScore	& 48.77 &	49.99	&51.76&52.04\\
			\hline
			\multicolumn{2}{c|}{Training time (h/epoch)}  & 46.99  & 52.12  & 55.36  & 59.10\\
			\hline
		\end{tabular}
	}
	\caption{Ablation study of routing expert numbers}
	\label{ablation_expert_number}
\end{table}
The number of experts represents a critical hyperparameter in EchoVLM where prior studies have demonstrated that scaling this parameter yields performance benefits. However its applicability within the ultrasound domain remains underexplored necessitating systematic investigation through ablation studies. As shown in Table~\ref{ablation_expert_number}, to address this gap we conducted an ablation experiment expanding the expert count from 0 to 6 specifically to determine whether the capacity gains observed in general VLMs could be effectively transferred to ultrasound-specific tasks characterized by distinct modality interactions and diagnostic complexity. Increasing from 0 to 4 experts delivers notable gains in high-complexity tasks report generation 10.15 BLEU-1 improvement ultrasound diagnosis 4.11 BLEU-1 improvement via improved model capacity enabling fine-grained cross-modal alignment. Compared to the 4-expert configuration six experts only yield marginal gains e.g. report generation BLEU-1 53.87 to 54.03 ultrasound diagnosis BLEU-1 62.81 to 63.23 but drastically increase training time 4 experts 55.36 h/epoch 6 experts 59.10 h/epoch. In contrast the 4-expert configuration achieves an optimal trade-off balancing superior performance and feasible training costs. 

\section{Conclusion}
This study introduces EchoVLM, the first open-source 10B-parameter universal ultrasound-specialized VLM, with three core contributions: (1) curation of the largest multi-organ ultrasound dataset (208,941 clinical cases, 1.47M images across 7 anatomical systems); (2) development of a novel expert-validated few-shot prompting mechanism enabling a robust multi-task instruction fine-tuning pipeline; (3) integration of a Dual-path MoE architecture with dynamic routing, significantly enhancing adaptability to ultrasound’s heterogeneous imaging features. Experiments confirm EchoVLM outperforms SOTA across multiple anatomical systems. Limitations include dataset long-tail distribution and complex visual question answering requiring multi-step reasoning. Future work will focus on data rebalancing, expanded anatomical coverage, longitudinal patient data integration, and routing refinement to advance precise, efficient AI-assisted ultrasound diagnostics.

\section{Limitations}
While EchoVLM achieves promising overall performance, it exhibits suboptimal performance in vascular analysis. We hypothesize that this limitation stems from the long-tailed distribution of the entire dataset, where vascular cases account for the smallest proportion. This data scarcity likely led domain experts to prioritize dominant anatomical patterns from majority classes (e.g., breast/liver), inadvertently marginalizing subtle vascular features. Collectively, these findings demonstrate EchoVLM's robustness in analyzing prevalent anatomical structures, while underscoring that its performance in vascular analysis could be significantly enhanced through data rebalancing strategies. Additionally, in open-ended tasks such as VQA, EchoVLM underperforms Lingshu (7B) in certain metrics, indicating potential room for improvement in its ability to handle the unstructured, open-ended nature of VQA tasks compared to general medical VLMs.
\section{Acknowledgments}
This work was supported by the National Natural Science Foundation of China (NSFC) Joint Fund for Regional Innovation and Development (Grant No. U25A20448), the National Natural Science Foundation of China (NSFC) Tianyuan Fund for Mathematics (Grant No. 12326609), the Guangzhou Key Research and Development Program (Grant No. 2025B03J0125), the Guangzhou Science and Technology Program - Key Research and Development (Grant No.: 2025B03J0155), and the National Natural Science Foundation of China (NSFC) General Program (Grant No.: 82572323). 

\bibliography{custom}
\clearpage
\appendix
\section{Implementation of Data Generation Pipeline: Desensitization and Quality Control}
\label{sec:appendix_a}
Data were collected from 15 hospitals across China, encompassing The First Affiliated Hospital of Sun Yat-sen University, The Third Affiliated Hospital of Sun Yat-sen University, The Sixth Affiliated Hospital of Sun Yat-sen University, The Seventh Affiliated Hospital of Sun Yat-sen University, The Eighth Affiliated Hospital of Sun Yat-sen University, The First Affiliated Hospital of Guangxi Medical University, The First Affiliated Hospital of Guangzhou Medical University, Sanshui District People's Hospital of Foshan City, Puer City People's Hospital, Sun Yat-sen University Cancer Center, Guangzhou Women and Children's Health Hospital, Guangzhou First People's Hospital, The First Affiliated Hospital of Guangzhou University of Chinese Medicine, Guangzhou Military Region General Hospital, and West China Hospital. Given the retrospective nature of this study, all collected data were de-identified prior to the initiation of model training, thereby waiving the requirement for informed consent. Both the data collection protocol and its subsequent utilization were approved by the appropriate institutional ethics review board (IRB).
\begin{figure}[ht]
	\centering
	\includegraphics[width=\linewidth]{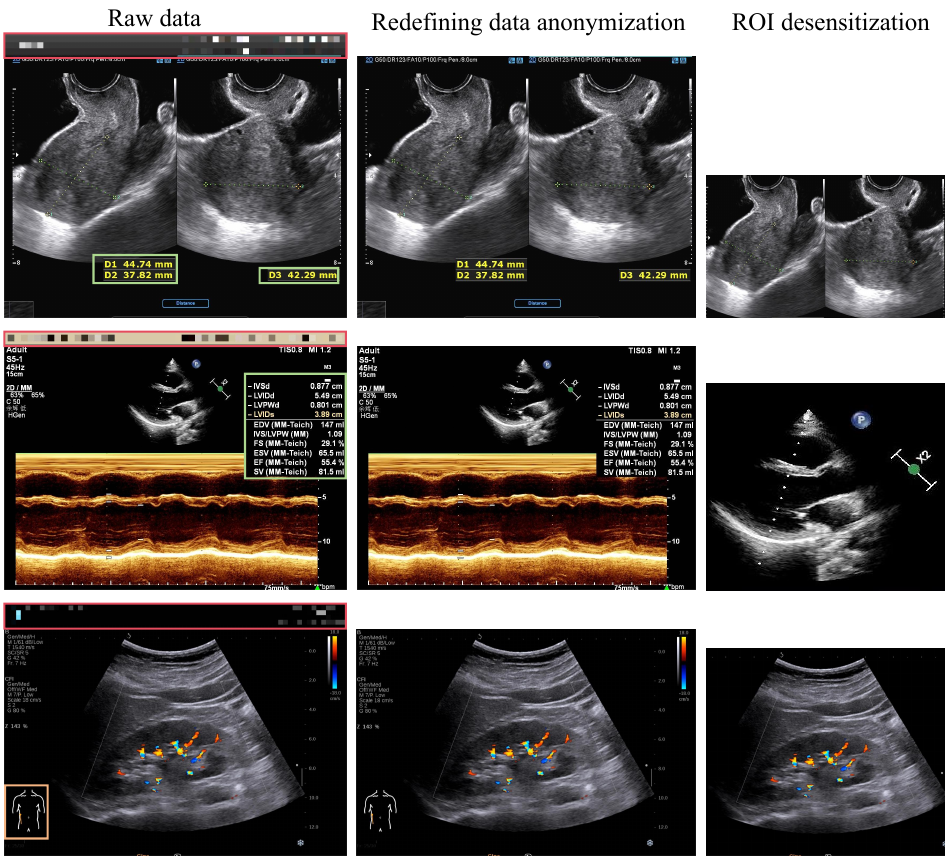}
	\caption{Comparison of traditional and redefined ROI desensitization.}
	\label{fig_4}
\end{figure}

As illustrated in Figure \ref{fig_4}, a comparative analysis was conducted between traditional ROI-based image data desensitization and the redefined image data desensitization approach proposed in this study. A core design objective of our redefined approach is to deliberately retain two types of critical information in the data: meaningful measurement values and anatomical information indicated by anatomical landmarks. Such retention is highly beneficial for improving the performance and clinical applicability of subsequent model training. Specifically, in the leftmost column of raw data, mosaic regions marked with red boxes correspond to sensitive information that needs to be desensitized; green box regions represent the measurement values we aim to preserve, which provide quantitative features for the model and assist it in capturing key numerical information critical for generating accurate and detailed clinical reports; yellow box regions denote the anatomical landmark information we intend to retain, which helps the model accurately locate and understand anatomical regions, thereby enhancing the clinical relevance and interpretability of the model’s outputs.
\begin{figure}[t]
	\centering
	\includegraphics[width=\linewidth]{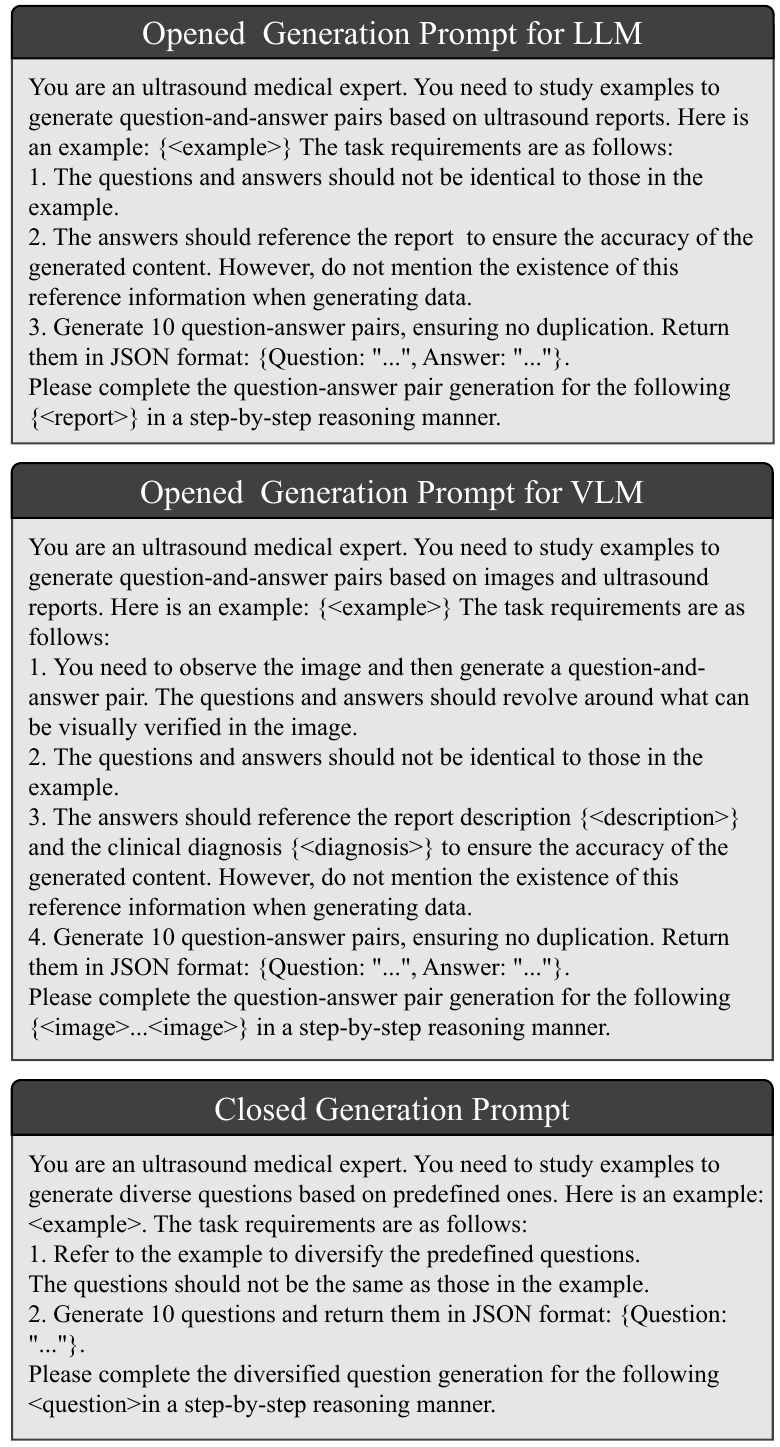}
	\caption{Prompt Templates for Ultrasound Q\&A Generation.}
	\label{fig_5}
\end{figure}

The data generation pipeline adopts a few-shot prompting strategy to produce high-quality training data for the model, with the utilized opened prompts and closed prompts presented in Figures \ref{fig_5}. A diverse array of advanced language and vision-language models was employed for data generation, including GPT-3.5-turbo, GPT-4V, Qwen2-VL-72B, Qwen2.5-VL-7B, Qwen2.5-VL-72B, Llama-3.1-8B, and Llama-3.1-70B. Such diversity in model selection is crucial for enhancing the generalization capability of the subsequent model training process, as it mitigates the risk of overfitting to biased or limited data distributions.
\begin{figure}[ht]
	\centering
	\includegraphics[width=\linewidth]{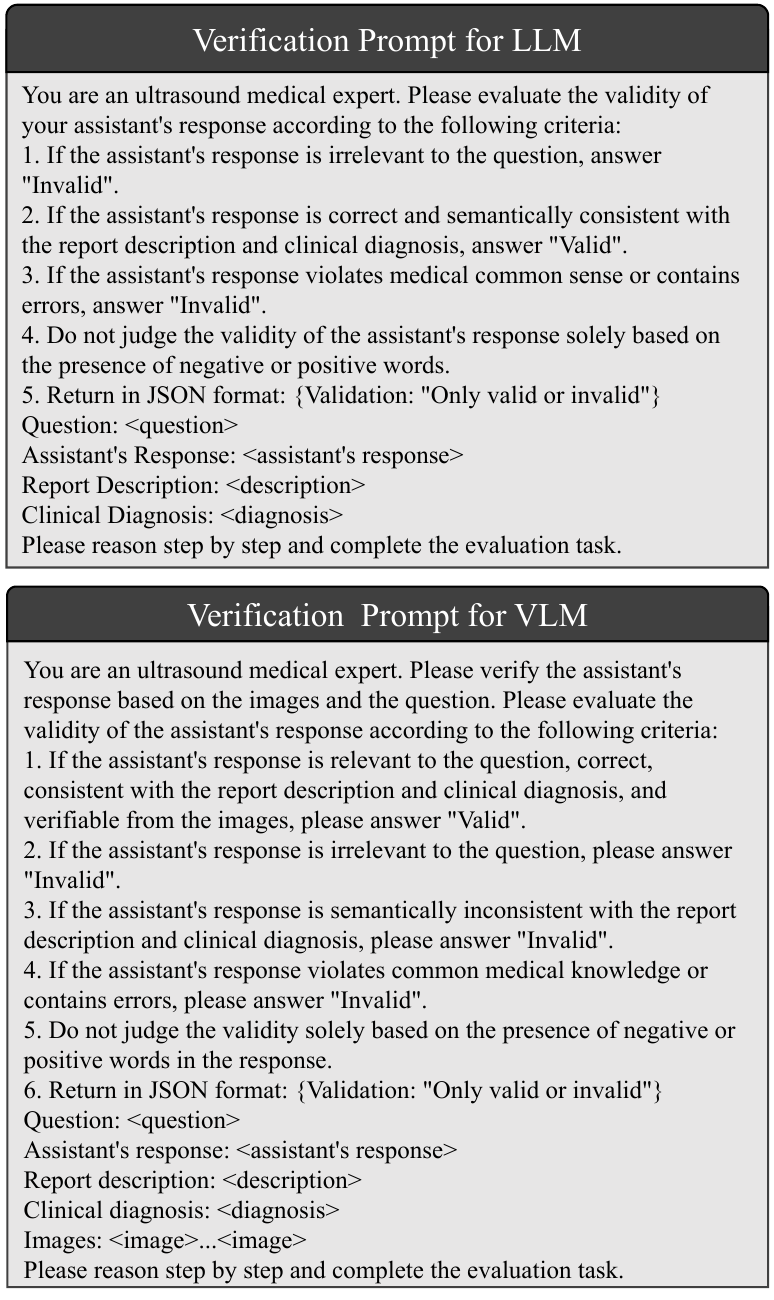}
	\caption{Ultrasound Q\&A verification prompt template.}
	\label{fig_6}
\end{figure}
Furthermore, strict quality control is integrated into the pipeline to ensure data reliability and clinical validity, with rigorous deduplication during instance creation to preserve content diversity. A newly generated instance is discarded if its ROUGE-L similarity with any previously generated instance exceeds 0.7, or if its Simhash similarity which is quantified by a Hamming distance threshold of $\le 3$ indicates substantial content overlap. For the open-ended data subset, an additional quality control layer is introduced, employing a dual-validation mechanism consisting of sequential AI pre-screening followed by expert verification. The prompts used for AI filtering are illustrated in Figure \ref{fig_6}. 
 Following the initial AI pre-screening step, 40,000 samples are randomly selected for clinical validation by domain experts. If errors, inconsistencies, or clinically inappropriate samples are identified in a selected sample set, the generation strategy or the source model associated with that sample set undergoes retrospective analysis and necessary adjustments, including model replacement and prompt optimization. This entire quality control process ultimately reduces 3.42 million raw instructions to 1.8 million high-quality data pairs.
\begin{figure*}[!ht]
	\subfloat[Model architecture and training framework]{
		\includegraphics[width=0.66\linewidth]{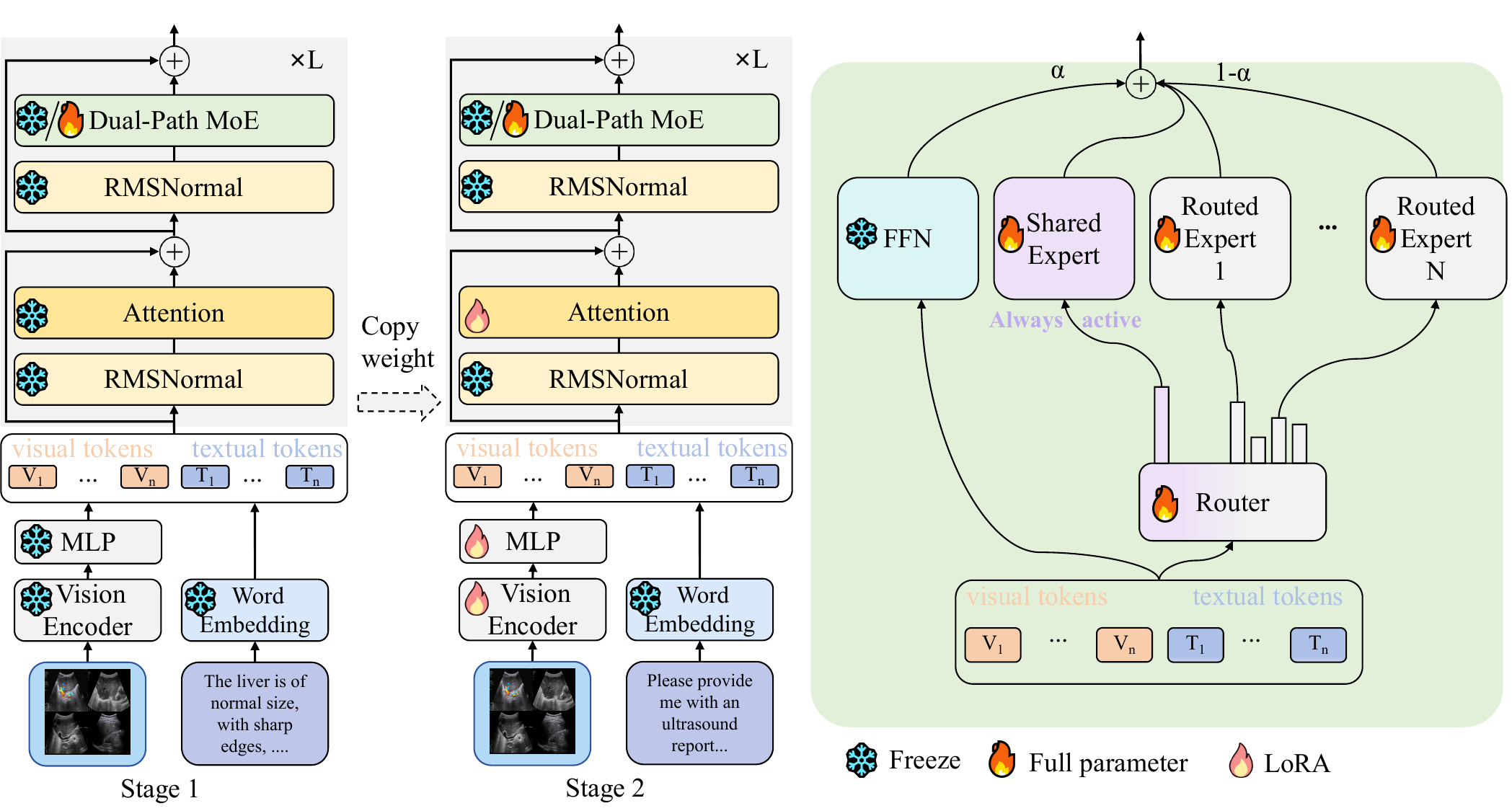}} \hfill
	\subfloat[Overall performance]{
		\includegraphics[width=0.34\linewidth]{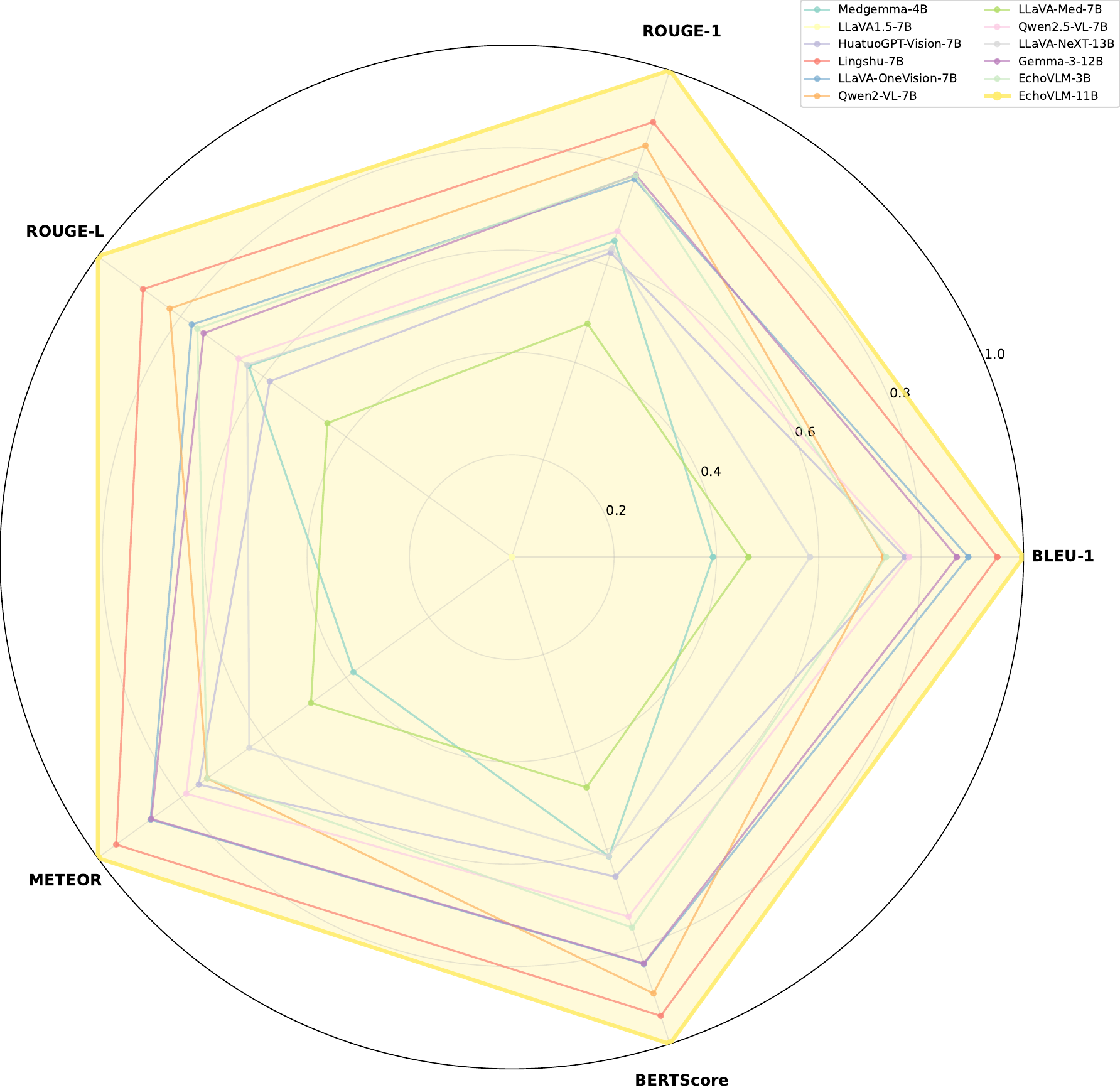}}
	\caption {Model structure, training strategy, and overall performance.}
	\label{model_architecture_and_performance}
\end{figure*}
\begin{table*}[!ht]
	\centering
	\small
	\resizebox{\linewidth}{!}{
		\begin{tabular}{llcccccccccccc}
			\toprule
			\textbf{Anatomical} & \textbf{Metric} & \textbf{Medgemma} & \textbf{LLaVA1.5} & \textbf{HuatuoGPT-Vision} & \textbf{Lingshu} & \textbf{LLaVA-OneVision} & \textbf{Qwen2-VL} & \textbf{LLaVA-Med} & \textbf{Qwen2.5-VL} & \textbf{LLaVA-NeXT} & \textbf{Gemma-3} & \textbf{EchoVLM} & \textbf{EchoVLM} \\
			\textbf{System} & & \textbf{(4B)} & \textbf{(7B)} & \textbf{(7B)} & \textbf{(7B)} & \textbf{(7B)} & \textbf{(7B)} & \textbf{(7B)} & \textbf{(7B)} & \textbf{(13B)} & \textbf{(12B)} & \textbf{(3B)} & \textbf{(11B)} \\
			\midrule
			\multirow{4}{*}{Breast} 
			& Hamming Accuracy & 61.22 & 70.49 & 72.81 & 73.03 & 72.78 & \cellcolor{second}74.08 & 73.38 & \cellcolor{third}73.54 & 72.44 & 69.59 & 73.30 & \cellcolor{first}\textbf{75.27} \\
			& Macro-Precision & 51.09 & 51.09 & \cellcolor{first}\textbf{62.97} & \cellcolor{third}62.11 & 61.03 & \cellcolor{second}62.30 & 55.77 & 55.47 & 61.59 & 59.70 & 48.23 & 59.24 \\
			& Macro-Recall & 30.81 & 49.37 & \cellcolor{first}\textbf{59.29} & 57.27 & \cellcolor{second}58.60 & 57.22 & 53.05 & \cellcolor{third}58.51 & 52.39 & 57.66 & 53.92 & 57.90 \\
			& Macro-F1 & 35.44 & 48.07 & \cellcolor{first}\textbf{58.27} & 56.31 & \cellcolor{second}57.16 & 54.24 & 49.68 & 54.55 & 50.54 & \cellcolor{second}57.16 & 49.18 & 54.37 \\
			\midrule
			\multirow{4}{*}{Gynecology}
			& Hamming Accuracy & \cellcolor{second}76.42 & 73.75 & 75.53 & 75.71 & 75.20 & 75.67 & 75.67 & 72.79 & 75.82 & 74.24 & \cellcolor{third}76.38 & \cellcolor{first}\textbf{78.28} \\
			& Macro-Precision & 48.98 & \cellcolor{second}60.88 & 45.56 & 47.52 & 46.27 & 42.57 & 44.05 & 46.06 & \cellcolor{first}\textbf{65.36} & 48.56 & 40.74 & \cellcolor{third}56.94 \\
			& Macro-Recall & 51.13 & 44.01 & 52.36 & \cellcolor{second}54.25 & \cellcolor{third}53.55 & 46.63 & 43.99 & 48.70 & 44.32 & 49.85 & 43.49 & \cellcolor{first}\textbf{56.18} \\
			& Macro-F1 & 47.72 & 41.67 & 47.15 & \cellcolor{second}48.71 & \cellcolor{third}48.22 & 42.13 & 39.62 & 44.36 & 40.46 & 47.80 & 39.60 & \cellcolor{first}\textbf{52.92} \\
			\midrule
			\multirow{4}{*}{Heart}
			& Hamming Accuracy & 84.83 & 77.06 & 84.28 & \cellcolor{third}87.75 & 85.95 & 85.31 & 83.42 & 86.86 & 84.56 & \cellcolor{second}87.77 & 85.01 & \cellcolor{first}\textbf{87.84} \\
			& Macro-Precision & 75.35 & \cellcolor{third}86.80 & 81.95 & 83.76 & 80.17 & 76.70 & 74.02 & 78.40 & \cellcolor{first}\textbf{91.48} & 82.39 & 73.93 & \cellcolor{second}\textbf{88.60} \\
			& Macro-Recall & 73.24 & 67.21 & 78.35 & 77.85 & \cellcolor{first}\textbf{82.34} & 74.69 & 71.51 & \cellcolor{second}80.16 & 73.32 & 78.44 & 74.06 & \cellcolor{third}79.31 \\
			& Macro-F1 & 72.23 & 70.08 & 78.36 & 77.81 & \cellcolor{second}78.67 & 74.08 & 71.21 & \cellcolor{third}78.47 & 74.57 & 78.24 & 73.25 & \cellcolor{first}\textbf{80.78} \\
			\midrule
			\multirow{4}{*}{Kidney}
			& Hamming Accuracy & 78.93 & 72.81 & 75.98 & \cellcolor{third}79.01 & \cellcolor{second}79.58 & 74.64 & 76.28 & 74.62 & 76.66 & 77.92 & 75.24 & \cellcolor{first}\textbf{81.96} \\
			& Macro-Precision & 56.03 & 59.93 & 58.11 & \cellcolor{third}60.73 & 53.64 & 53.96 & 48.42 & 51.77 & \cellcolor{second}60.82 & 59.86 & 50.01 & \cellcolor{first}\textbf{65.60} \\
			& Macro-Recall & 44.53 & 38.52 & \cellcolor{third}48.32 & \cellcolor{second}51.51 & 43.94 & 38.07 & 35.38 & 38.95 & 36.46 & 46.78 & 35.04 & \cellcolor{first}\textbf{61.12} \\
			& Macro-F1 & 45.65 & 42.01 & \cellcolor{third}51.57 & \cellcolor{second}53.01 & 42.24 & 41.92 & 36.48 & 42.19 & 37.87 & 49.12 & 36.95 & \cellcolor{first}\textbf{61.61} \\
			\midrule
			\multirow{4}{*}{Liver}
			& Hamming Accuracy & \cellcolor{second}81.41 & 70.44 & 76.73 & 79.04 & 79.87 & 76.67 & 76.67 & 71.56 & 77.04 & 76.81 & \cellcolor{first}\textbf{82.14} & \cellcolor{third}81.19 \\
			& Macro-Precision & 51.96 & \cellcolor{third}53.90 & 50.24 & 50.66 & 51.98 & 48.56 & 52.91 & 48.53 & \cellcolor{first}\textbf{58.34} & 51.20 & 48.18 & \cellcolor{second}57.22 \\
			& Macro-Recall & 42.68 & 28.86 & 38.61 & \cellcolor{third}44.04 & \cellcolor{second}48.24 & 36.53 & 35.37 & 31.78 & 35.88 & 43.91 & 43.48 & \cellcolor{first}\textbf{49.77} \\
			& Macro-F1 & 42.32 & 34.55 & 41.77 & \cellcolor{third}45.96 & 45.65 & 39.62 & 37.83 & 36.74 & 38.16 & \cellcolor{second}46.13 & 42.65 & \cellcolor{first}\textbf{50.24} \\
			\midrule
			\multirow{4}{*}{Vessel}
			& Hamming Accuracy & 71.89 & 54.56 & 61.39 & \cellcolor{third}72.22 & 67.72 & 68.44 & 22.28 & 66.17 & 54.10 & \cellcolor{first}\textbf{73.06} & \cellcolor{second}72.39 & 51.83 \\
			& Macro-Precision & 72.85 & 71.28 & 79.08 & 80.22 & 78.65 & 79.18 & 13.61 & 79.04 & 69.71 & \cellcolor{third}81.70 & \cellcolor{second}87.20 & \cellcolor{first}\textbf{92.88} \\
			& Macro-Recall & \cellcolor{third}60.20 & 39.95 & 55.00 & \cellcolor{second}60.34 & 58.55 & 50.50 & 10.05 & 53.28 & 39.55 & \cellcolor{first}\textbf{64.10} & 56.98 & 33.85 \\
			& Macro-F1 & 63.17 & 47.94 & 60.31 & \cellcolor{second}63.86 & 62.86 & 55.59 & 4.55 & 60.35 & 46.99 & \cellcolor{first}\textbf{68.12} & \cellcolor{third}63.22 & 45.46 \\
			\midrule
			\multirow{4}{*}{Thyroid}
			& Hamming Accuracy & 68.81 & 65.85 & 75.02 & \cellcolor{third}78.50 & 74.21 & \cellcolor{second}78.88 & 77.45 & 76.27 & 76.98 & 73.85 & 76.80 & \cellcolor{first}\textbf{80.70} \\
			& Macro-Precision & 39.82 & 38.87 & 47.13 & 47.49 & 41.25 & \cellcolor{first}\textbf{54.63} & 37.06 & 46.82 & 36.95 & \cellcolor{third}49.42 & 44.29 & \cellcolor{second}54.44 \\
			& Macro-Recall & 22.74 & 20.25 & 36.23 & \cellcolor{second}42.31 & 34.87 & 39.05 & \cellcolor{third}41.41 & 38.75 & 41.32 & 34.50 & 40.19 & \cellcolor{first}\textbf{47.35} \\
			& Macro-F1 & 27.05 & 24.35 & 37.94 & \cellcolor{second}41.65 & 35.73 & 38.13 & 36.73 & 37.26 & 36.57 & \cellcolor{third}38.26 & 37.43 & \cellcolor{first}\textbf{46.12} \\
			\midrule
			\multirow{4}{*}{Average}
			& Hamming Accuracy & 74.79 & 69.28 & 74.53 & \cellcolor{first}\textbf{77.89} & 76.47 & 76.24 & 69.31 & 74.54 & 73.94 & 76.18 & \cellcolor{second}77.32 & \cellcolor{third}76.72 \\
			& Macro-Precision & 56.58 & 60.39 & 60.72 & 61.78 & 58.99 & 59.70 & 46.55 & 58.01 & \cellcolor{second}63.46 & \cellcolor{third}61.83 & 56.08 & \cellcolor{first}\textbf{67.85} \\
			& Macro-Recall & 46.48 & 41.17 & 52.59 & \cellcolor{first}\textbf{55.37} & \cellcolor{third}54.30 & 48.96 & 41.54 & 50.02 & 46.18 & 53.61 & 49.59 & \cellcolor{second}55.07 \\
			& Macro-F1 & 47.65 & 44.10 & 53.62 & \cellcolor{second}55.33 & 52.93 & 49.39 & 39.44 & 50.56 & 46.45 & \cellcolor{third}54.98 & 48.90 & \cellcolor{first}\textbf{55.93} \\
			\bottomrule
		\end{tabular}
	}
	\caption{Multi-label Entity Recognition Performance on Ultrasound Reports.}
	\label{tab:report_entity}
\end{table*}
\section{Model Architecture and Training Configuration}
\label{sec:appendix_b}
As presented in Table \ref{table_7}, we detail the model architecture and our proposed two-stage training framework, which integrates a Dual-path MoE for incremental training to enhance domain-specific adaptation. Specifically, the routing experts within this MoE are configured with a dimensionality of 1408, adhering to the Qwen model specifications, while employing a Top-2 selection strategy among four experts to balance specialization and efficiency; concurrently, the shared expert dimension is established at four times that of the routing experts to ensure sufficient capacity for acquiring comprehensive general ultrasound knowledge. Additionally, the PatchMerger module facilitates the fusion and compression of visual representations by consolidating adjacent 2×2 tokens into a single token, thereby optimizing visual feature processing. As shown in Figure \ref{model_architecture_and_performance}, during Stage I, the base model parameters are maintained in a frozen state to preserve pre-trained representations, with optimization strictly confined to the newly integrated Dual-path MoE components for acquiring domain-specific visual-textual patterns. In the subsequent Stage II, Cooperative Instruction Fine-tuning is implemented via Low-Rank Adaptation (LoRA), selectively updating the visual encoder, vision-to-text projection layer, and large language model attention layers, while concurrently enabling full-parameter updates for the active experts within the Dual-path MoE to maintain their domain-specialized adaptation capabilities. All baseline models in both comparative and preliminary experiments follow the full-parameter fine-tuning paradigm of LLaVA, which ensures comparable parameter updates with EchoVLM across both training stages and enables a fair and rigorous experimental comparison.
\section{Routing Distributions}
\begin{figure*}[t]
	\centering
	\resizebox{0.7\textwidth}{!}{
		\parbox{\textwidth}{
	\subfloat[Breast]{%
		\includegraphics[width=\linewidth]{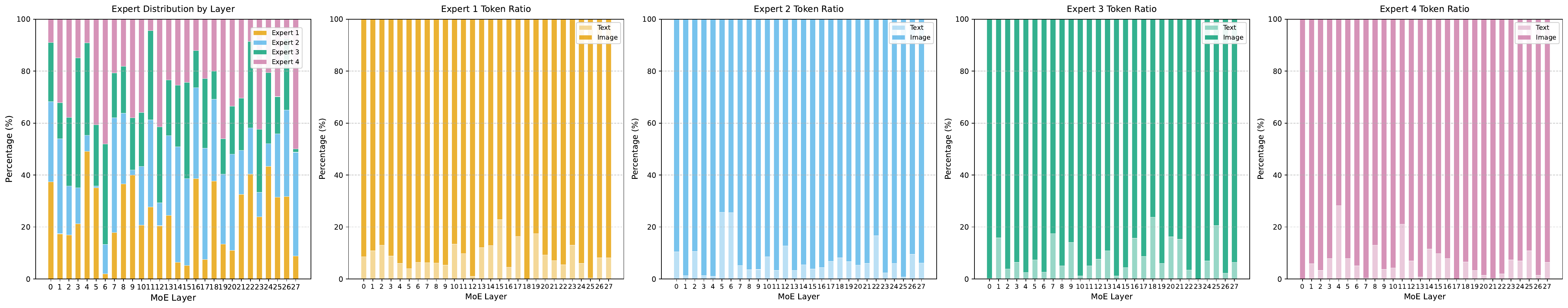}}\hfill
	\subfloat[Gynaecology]{%
		\includegraphics[width=\linewidth]{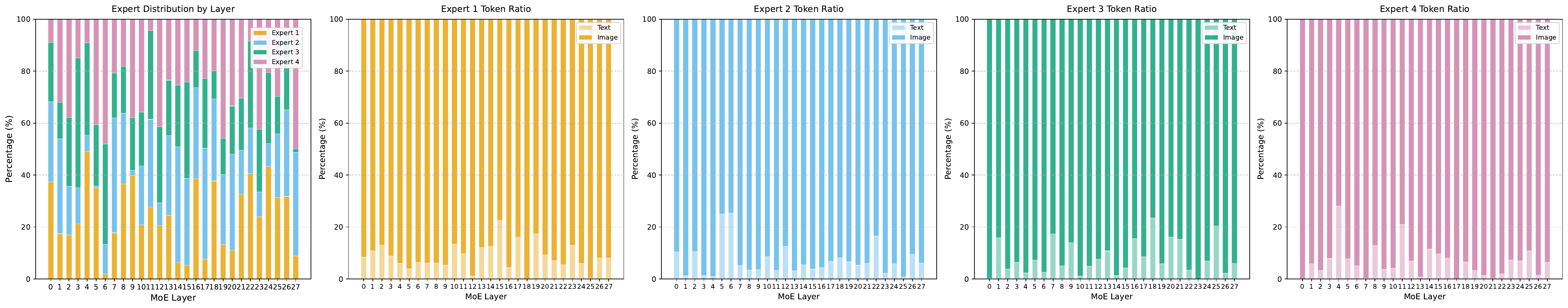}}\hfill
	\subfloat[Heart]{%
		\includegraphics[width=\textwidth]{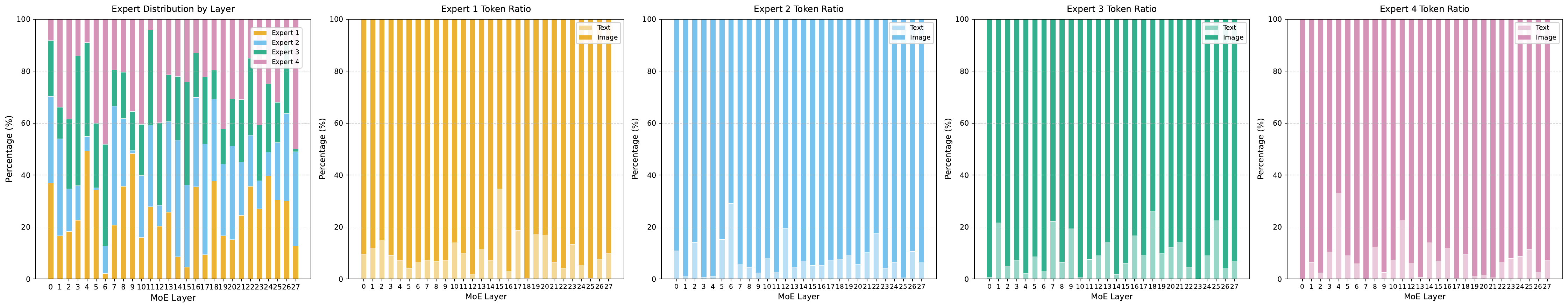}}\\
	\subfloat[Kidney]{%
		\includegraphics[width=\textwidth]{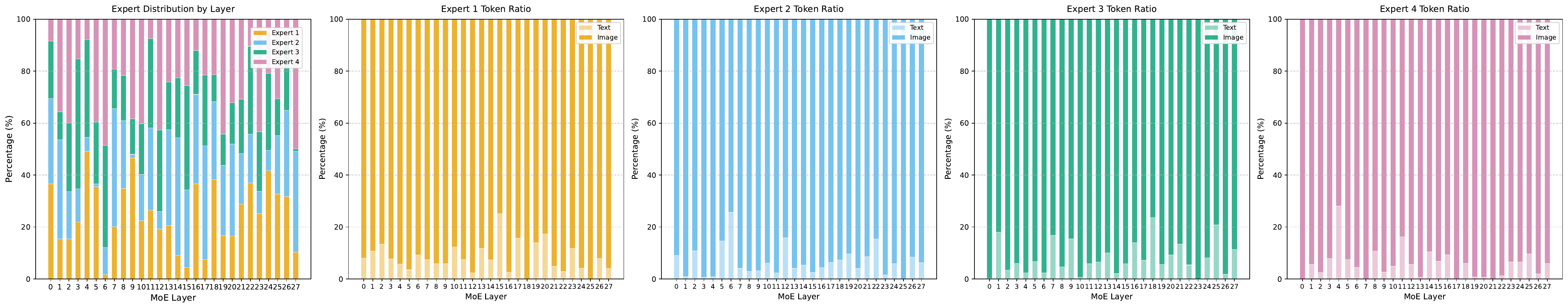}}\\
	\subfloat[Liver]{%
		\includegraphics[width=\textwidth]{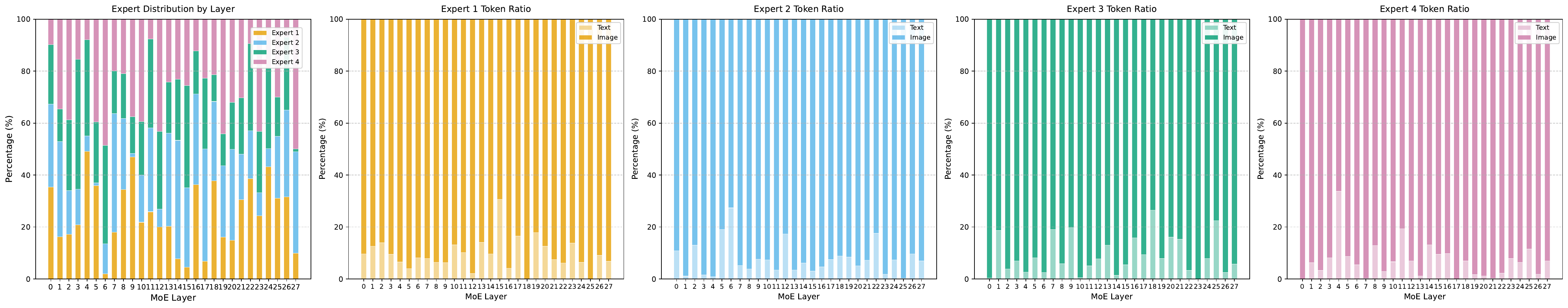}}\\
	\subfloat[Thyroid]{%
		\includegraphics[width=\textwidth]{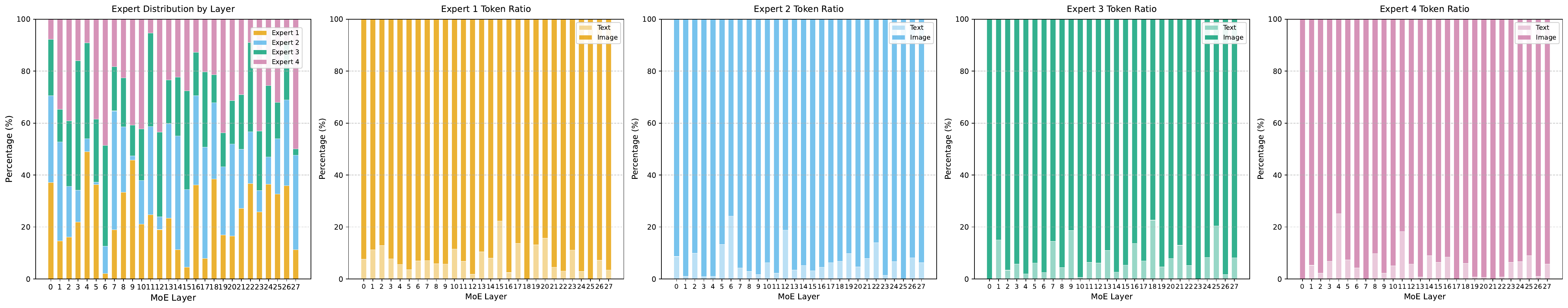}}\\
	\subfloat[Vessel]{%
		\includegraphics[width=\textwidth]{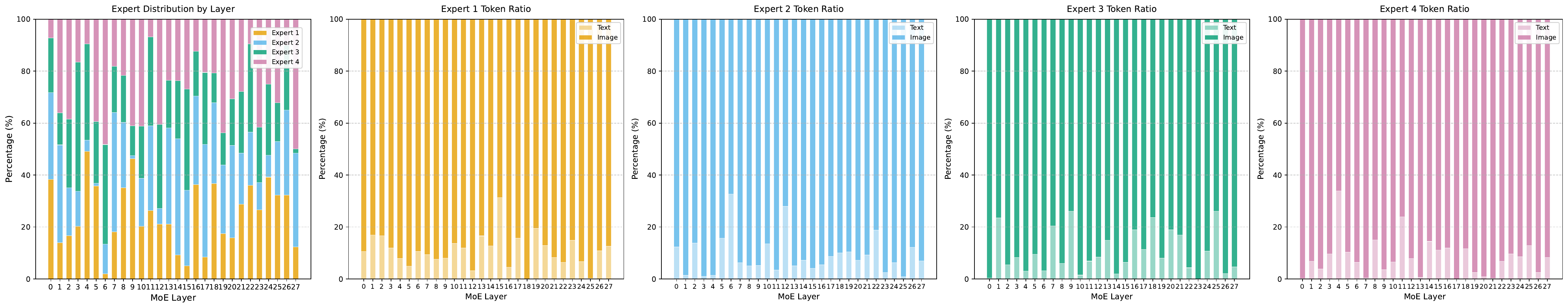}}\\
	\subfloat[Overall]{%
		\includegraphics[width=\textwidth]{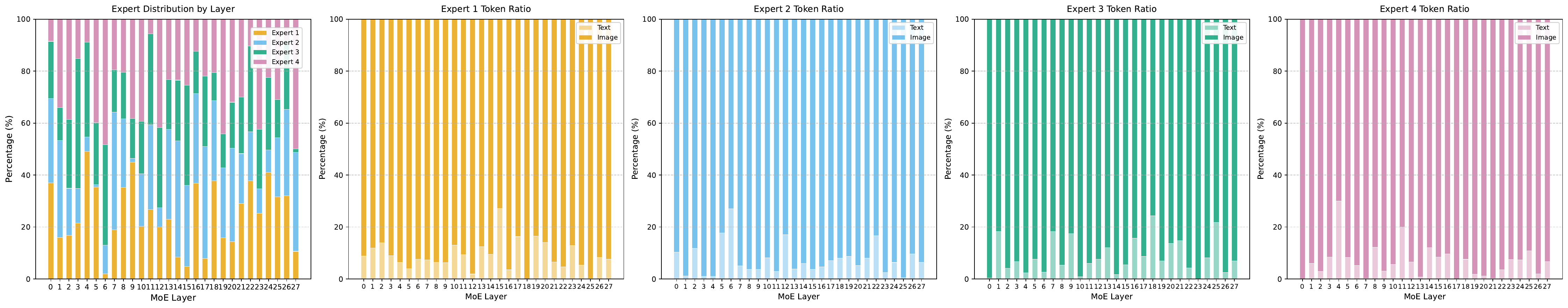}}
	}
	}
	\caption{Expert routing distribution analysis for different medical imaging tasks: 
		(a) Breast, (b) Gynaecology, (c) Heart, (d) Kidney, (e) Liver, (f) Thyroid, (g) Vessel, and (h) Overall}
	\label{expert_distribution}
\end{figure*}

As shown in Figure \ref{expert_distribution}, we conduct a comprehensive analysis of the routing distribution within the MoE framework on the test set to rigorously evaluate its efficacy and load-balancing performance. As illustrated in the accompanying figure, the aggregate routing frequencies across all experts exhibit a remarkably balanced pattern, demonstrating that no single expert is disproportionately over- or under-utilized.
Delving deeper into modality-specific routing behavior, we observe that the overwhelming majority of dispatched tokens correspond to image data, whereas textual tokens constitute a markedly smaller fraction. This pronounced imbalance originates from an inherent bias in our dataset composition: although we possess 208,941 radiological reports, these are accompanied by 1.47 million key-frame ultrasound images. Consequently, each individual report is associated with multiple images, thereby skewing the token distribution toward visual content and explaining the dominant image-centric routing observed across experts.
\section{Visualization}
\begin{figure*}[t]
	\centering
	\resizebox{0.6\textwidth}{!}{
	\parbox{\textwidth}{
	\subfloat{\includegraphics[width=1.0\textwidth]{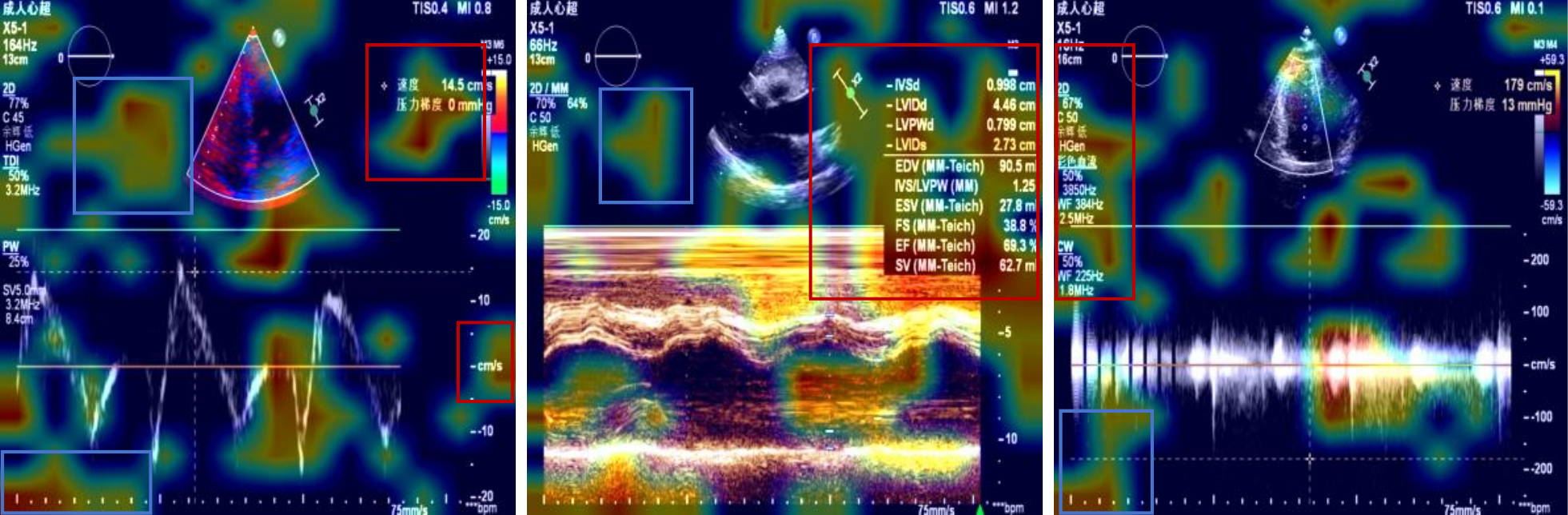}}\\
	\subfloat{\includegraphics[width=1.0\textwidth]{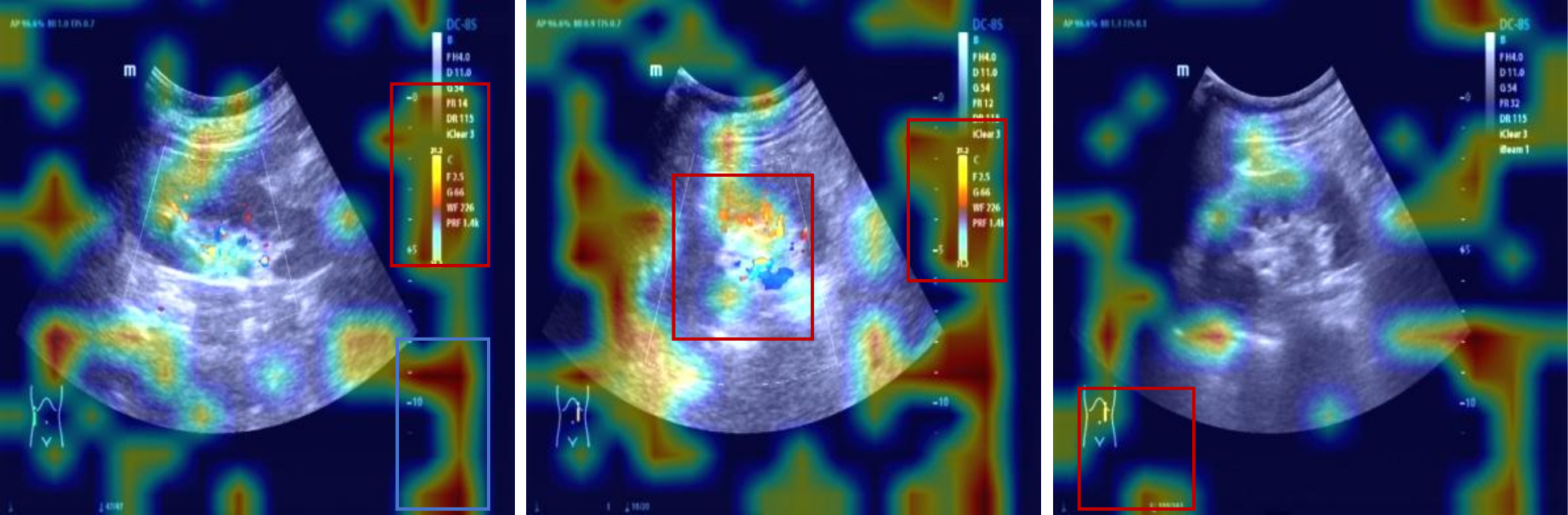}}\\
	\subfloat{\includegraphics[width=1.0\textwidth]{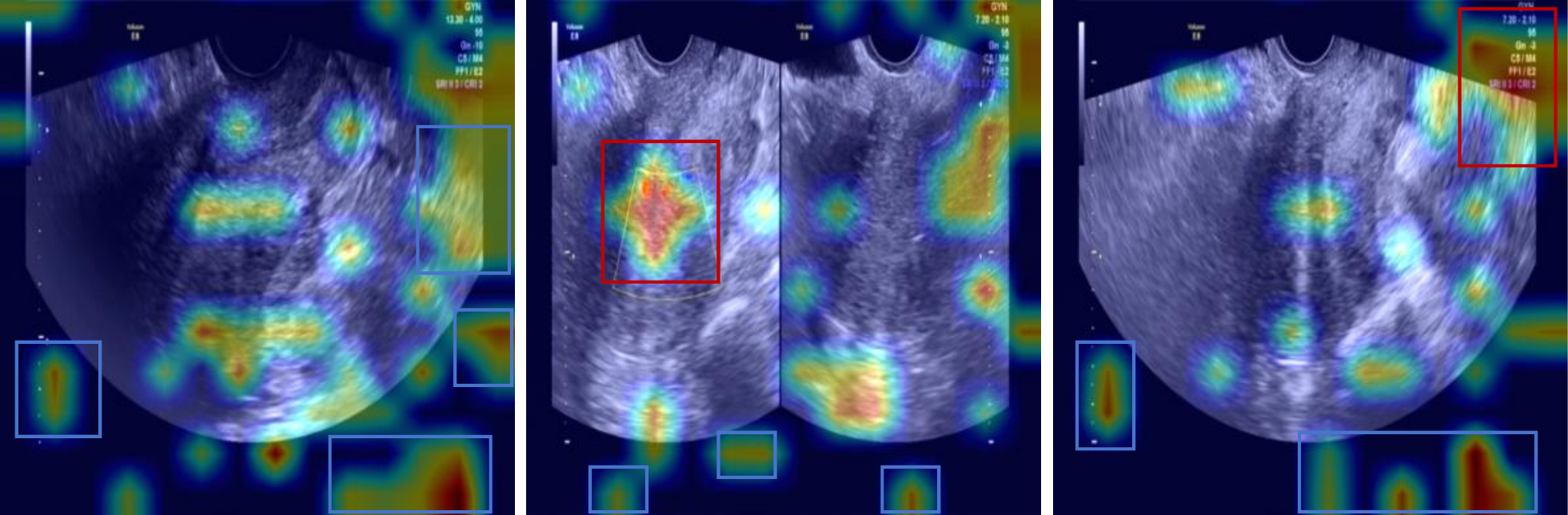}}\\
	\subfloat{\includegraphics[width=1.0\textwidth]{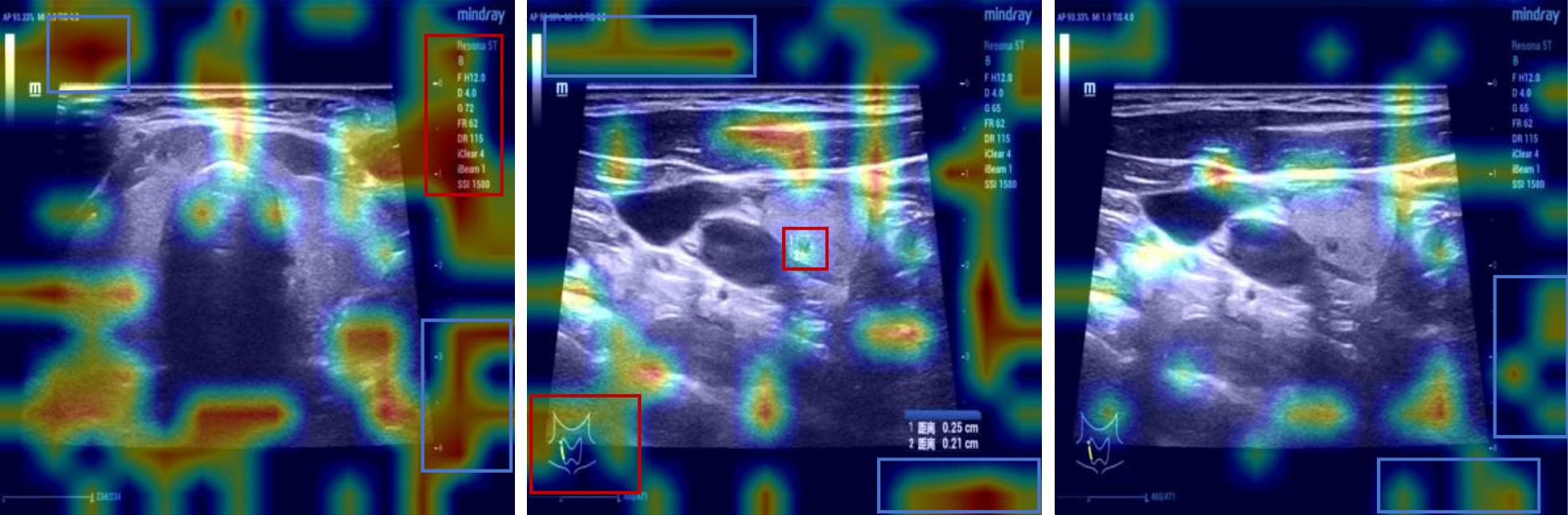}}\\
	\subfloat{\includegraphics[width=1.0\textwidth]{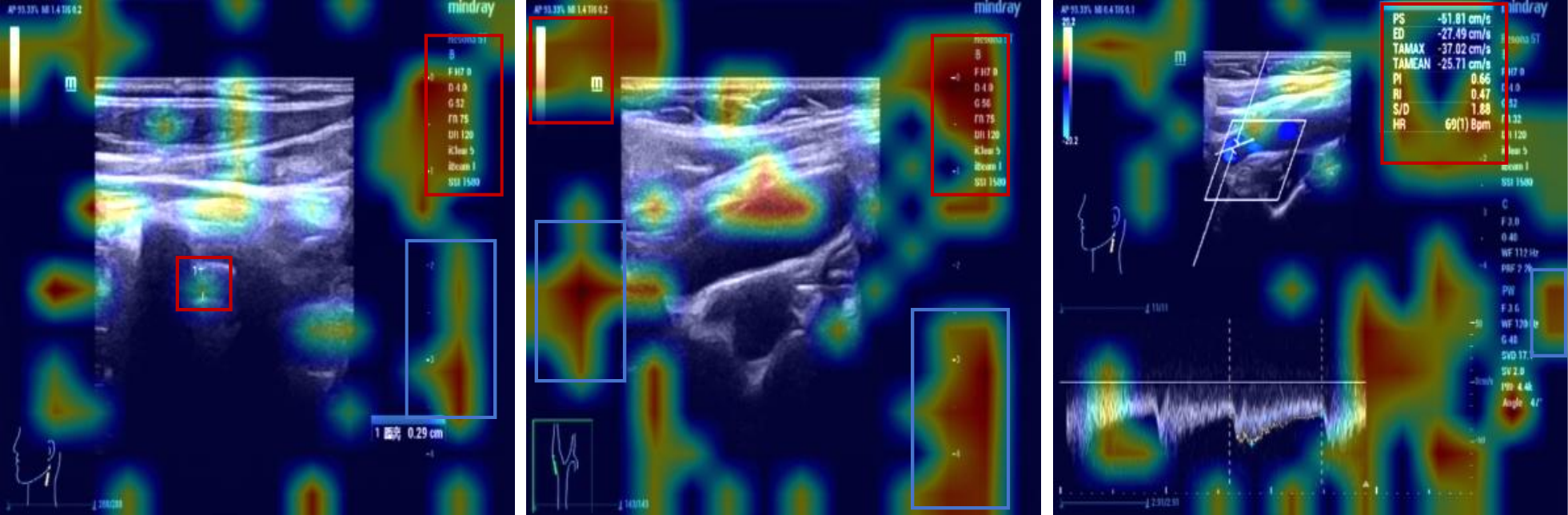}}\\
	\subfloat{\includegraphics[width=1.0\textwidth]{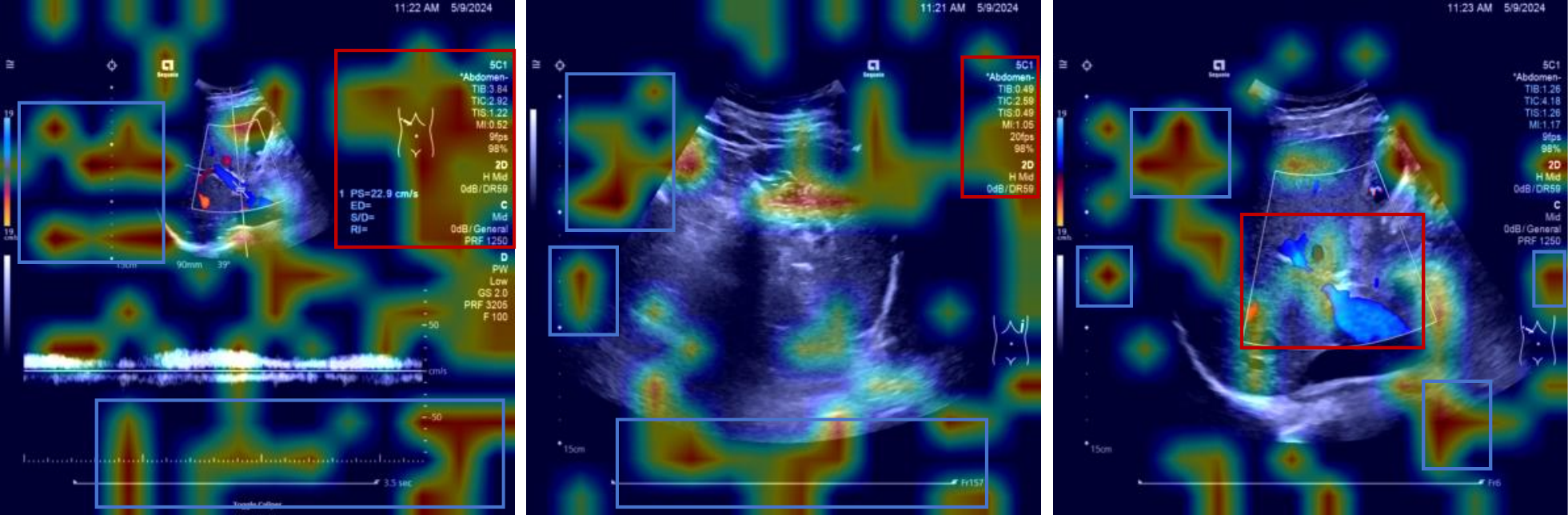}}\\
	\subfloat{\includegraphics[width=1.0\textwidth]{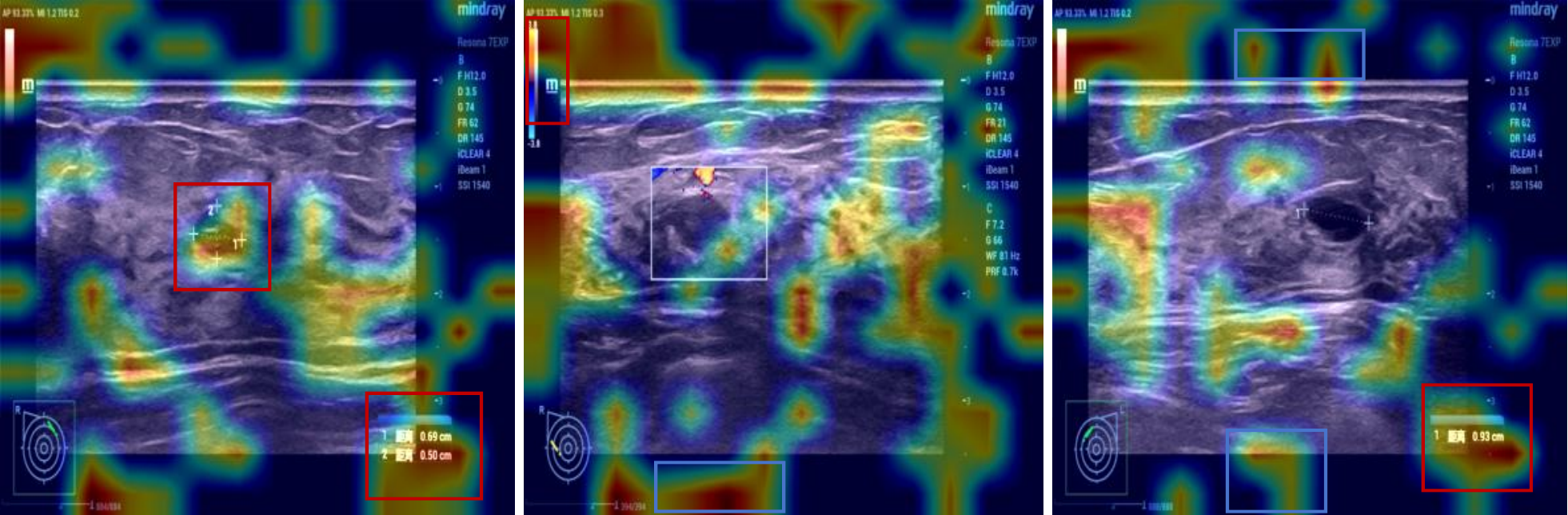}}
		}
}
	\caption{Visualization results. Red boxes highlight meaningful regions, while blue boxes indicate irrelevant or meaningless areas.}
		\label{Visualization}
\end{figure*}

To elucidate the internal decision-making process of VLMs, we employed the Grad-CAM heatmap visualization technique~\cite{cam}. In Figure \ref{Visualization}, warmer regions (depicted in red) indicate pixels with the strongest gradient flow towards the predicted token logit, signifying that the model primarily attends to these semantically significant regions during textual output generation. Cooler regions (in blue) receive negligible weight allocation, implying minimal contribution to the prediction. The visualized attention patterns confirm that the VLM's language reasoning is grounded in semantically relevant visual features such as anatomical landmarks, quantitative measurements, and hemodynamic bar charts, as highlighted within red bounding boxes, rather than spurious correlations. However, the model also exhibits attention to non-informative regions including image edges and black areas, as highlighted within blue bounding boxes, which lack semantic utility. This observation highlights a persistent challenge in VLMs: the coexistence of semantic consistency (focusing on task-critical visual elements) and spurious associations (distracting attention to noise or irrelevant artifacts). Future work should enhance model robustness and interpretability through strategies such as attention regularization, constrained attention routing, and explicit mitigation of spurious associations.
\begin{table}[ht]
	\small 
	\centering
		\begin{tabular}{l|c|c}
			\hline
			Config & Stage I & Stage II  \\ \hline
			Image encoder & \multicolumn{2}{c}{CLIP-ViT-L/14}  \\ \hline
			Vision-to-text projection  & \multicolumn{2}{c}{MLP}  \\ \hline
			LLM  & \multicolumn{2}{c}{Qwen2-2B/Qwen2-7B}  \\ \hline
			PatchMerger rate  & \multicolumn{2}{c}{4}  \\ \hline
			Shared expert number  & \multicolumn{2}{c}{1}  \\ \hline
			Shared expert dimension  & \multicolumn{2}{c}{1536/5632}  \\ \hline
			Routing expert number  & \multicolumn{2}{c}{4}  \\ \hline
			Routing expert dimension  & \multicolumn{2}{c}{768/1408}  \\ \hline
			Top-k   & \multicolumn{2}{c}{2}  \\ \hline
			Feature select layer & \multicolumn{2}{c}{-1}  \\ \hline
			Image resolution & \multicolumn{2}{c}{392×392}  \\ \hline
			LoRA rank & - &8  \\ \hline
			LoRA alpha & - &16  \\ \hline
			LoRA dropout & - &0.05  \\ \hline
			Deepspeed & \multicolumn{2}{c}{Zero2}  \\ \hline
			Epoch & \multicolumn{2}{c}{1}  \\ \hline
			Optimizer & \multicolumn{2}{c}{AdamW}  \\ \hline
			Weight decay & \multicolumn{2}{c}{0.0}  \\ \hline
			Learning rate &1e-3 &2e-5  \\ \hline
			Learning rate schdule & \multicolumn{2}{c}{Cosine}  \\ \hline
			Warmup ratio & \multicolumn{2}{c}{0.03}  \\ \hline
			Max length & \multicolumn{2}{c}{32768}  \\ \hline
			Batch size per GPU & \multicolumn{2}{c}{1}  \\ \hline
			GPU & \multicolumn{2}{c}{8×A100-80G}  \\ \hline
			Gradient checkpointing & \multicolumn{2}{c}{True}  \\ \hline
			Precision & \multicolumn{2}{c}{Bf16}  \\ \hline
			Auxiliary loss weight $\gamma$ & \multicolumn{2}{c}{0.001}  \\ \hline
			Training parameters  &1.38B/3.39B  &1.39B/3.4B  \\ \hline
			Total parameters  &  \multicolumn{2}{c}{3B/11B}  \\ \hline
		\end{tabular}
		\caption{Model Architecture and Training Configuration.}
		\label{table_7}
	\end{table}
	\begin{table*}[!ht]
		\small 
		\centering
		\resizebox{\linewidth}{!}{%
			\begin{tabular}{c|c|cccccccccccc}
				\hline
				Anatomical	&Metric$\uparrow$&\makecell{Medgemma\\(4B)}&\makecell{LLaVA1.5\\(7B)}&\makecell{HuatuoGPT-Vision\\(7B)}& \makecell{Lingshu\\(7B)} & \makecell{LLaVA-OneVision-\\(7B)}&\makecell{Qwen2-VL-\\(7B)}&\makecell{LLaVA-Med (7B)}&\makecell{Qwen2.5-VL\\(7B)}&\makecell{LLaVa-NeXT\\(13B)}&\makecell{Gemma-3\\(12B)}&\makecell{EchoVLM\\(3B)} &EchoVLM(11B) \\  
				\hline
				\multirow{4}{*}{Breast} 
				&Hamming Accuracy		&82.87&79.31&81.75&86.38&82.75&83.62&\cellcolor{third}86.81&86.31&86.34&\cellcolor{second}87.81&\cellcolor{first}\textbf{88.25}&\cellcolor{third}87.94\\ 
				&Macro-Precision 	&53.91&48.64&48.12&51.88&44.78&52.85&56.07&\cellcolor{second}65.01&55.25&54.40&55.63&\cellcolor{first}\textbf{65.43}\\
				&Macro-Recall  	&43.03&42.22&41.48&48.52&45.77&47.56&\cellcolor{third}55.07&51.30&\cellcolor{second}54.45&56.20&\cellcolor{first}\textbf{57.73}&\cellcolor{first}\textbf{58.11}\\ 
				&Macro-F1  	&43.63&37.68&41.70&47.39&42.35&46.82&53.29&48.68&\cellcolor{third}52.70&\cellcolor{second}55.11&\cellcolor{second}54.91&\cellcolor{first}\textbf{57.88}\\
				\hline
				\multirow{4}{*}{Gynecology} 
				&Hamming Accuracy		&88.39&87.42&88.53&87.38&86.86&89.28&\cellcolor{third}90.51&87.11&\cellcolor{second}90.59&89.22&\cellcolor{second}90.81&\cellcolor{first}\textbf{92.52}\\
				&Macro-Precision  	&46.40&\cellcolor{third}62.70&53.17&50.64&48.40&53.52&54.39&44.11&54.63&\cellcolor{second}58.73&\cellcolor{second}59.48&\cellcolor{first}\textbf{63.53}\\
				&Macro-Recall  	&29.79&22.58&44.03&42.68&41.08&46.02&39.27&43.63&39.42&38.51&\cellcolor{third}50.90&\cellcolor{first}\textbf{64.28}\\ 
				&Macro-F1  	&29.86&30.32&41.87&38.72&38.50&45.10&43.12&38.65&43.34&43.23&\cellcolor{third}48.90&\cellcolor{first}\textbf{61.75}\\
				\hline
				\multirow{4}{*}{Heart} 
				&Hamming Accuracy		&87.96&78.29&84.90&\cellcolor{third}90.83&87.67&\cellcolor{third}90.76&\cellcolor{second}93.57&87.90&\cellcolor{first}\textbf{93.82}&90.45&91.58&\cellcolor{second}92.71\\
				&Macro-Precision  	&57.67&45.12&57.63&70.57&63.84&66.47&\cellcolor{third}72.40&66.73&\cellcolor{first}\textbf{75.69}&65.59&65.13&\cellcolor{second}72.86\\
				&Macro-Recall  	&50.36&33.47&59.59&60.83&56.19&61.50&\cellcolor{third}65.91&51.52&\cellcolor{first}\textbf{67.57}&62.07&61.93&\cellcolor{second}69.90\\
				&Macro-F1  	&52.91&37.92&57.82&63.03&58.27&62.27&\cellcolor{third}66.40&54.95&\cellcolor{first}\textbf{68.47}&62.26&61.95&\cellcolor{second}69.93\\
				\hline
				\multirow{4}{*}{Kidney} 
				&Hamming Accuracy		&92.79&81.75&90.78&93.57&\cellcolor{third}94.60&93.73&93.71&92.25&93.71&\cellcolor{second}94.35&\cellcolor{first}\textbf{94.64}&\cellcolor{third}94.54\\
				&Macro-Precision  	&65.60&39.21&54.89&64.95&60.35&\cellcolor{third}69.22&\cellcolor{second}67.74&57.07&\cellcolor{second}67.73&\cellcolor{second}67.81&\cellcolor{first}\textbf{72.84}&\cellcolor{third}69.66\\
				&Macro-Recall  	&57.80&31.69&62.58&58.99&59.72&62.37&\cellcolor{second}65.45&58.48&\cellcolor{second}65.49&62.56&\cellcolor{third}64.18&\cellcolor{first}\textbf{73.58}\\
				&Macro-F1  	&59.11&32.69&56.57&58.33&59.15&63.72&\cellcolor{second}63.87&57.16&\cellcolor{first}\textbf{63.89}&62.51&\cellcolor{third}66.55&\cellcolor{first}\textbf{70.64}\\
				\hline
				\multirow{4}{*}{Liver} 
				&Hamming Accuracy		&92.28&88.09&92.83&93.48&93.50&93.50&92.53&92.57&92.65&90.92&\cellcolor{second}94.24&\cellcolor{first}\textbf{94.58}\\
				&Macro-Precision 	&39.63&33.09&47.30&47.35&49.63&47.11&45.85&47.21&\cellcolor{third}51.18&39.58&\cellcolor{second}56.50&\cellcolor{first}\textbf{64.65}\\
				&Macro-Recall  	&43.91&26.64&48.82&46.58&45.82&\cellcolor{third}47.87&\cellcolor{second}47.38&47.61&\cellcolor{third}48.16&44.33&46.04&\cellcolor{first}\textbf{58.09}\\
				&Macro-F1  	&40.71&26.52&45.16&44.28&46.04&44.74&43.00&43.22&44.33&38.50&\cellcolor{third}47.00&\cellcolor{first}\textbf{56.35}\\
				\hline
				\multirow{4}{*}{Vessel} 
				&Hamming Accuracy		&89.05&84.65&83.40&88.95&87.20&\cellcolor{third}90.30&\cellcolor{second}90.85&88.65&86.14&87.90&\cellcolor{first}\textbf{91.05}&88.20\\
				&Macro-Precision 	&44.13&32.85&27.12&45.50&45.33&\cellcolor{first}\textbf{63.67}&44.14&37.34&34.75&42.76&43.72&\cellcolor{second}51.06\\
				&Macro-Recall  	&31.71&28.07&34.30&31.46&\cellcolor{third}34.57&\cellcolor{first}\textbf{37.03}&31.38&29.72&30.67&29.96&32.07&27.19\\
				&Macro-F1  	&35.40&28.46&28.99&33.83&\cellcolor{second}36.75&\cellcolor{first}\textbf{44.49}&33.55&31.34&30.67&32.65&\cellcolor{third}36.11&34.26\\
				\hline
				\multirow{4}{*}{Thyroid} 
				&Hamming Accuracy		&76.96&69.36&77.31&76.56&79.02&79.14&75.18&78.29&74.87&77.00&\cellcolor{second}82.69&\cellcolor{first}\textbf{84.78}\\
				&Macro-Precision 	&63.17&54.79&65.49&68.54&66.82&\cellcolor{third}73.65&62.01&\cellcolor{second}68.92&61.57&62.81&\cellcolor{second}71.63&\cellcolor{first}\textbf{76.73}\\
				&Macro-Recall  	&57.75&33.91&61.29&59.32&\cellcolor{third}64.83&61.22&60.94&62.28&60.38&\cellcolor{second}62.97&\cellcolor{second}68.69&\cellcolor{first}\textbf{75.05}\\ 
				&Macro-F1  	&59.25&39.21&59.34&58.93&\cellcolor{third}64.70&60.11&59.69&\cellcolor{second}62.62&59.12&60.54&\cellcolor{second}68.72&\cellcolor{first}\textbf{71.21}\\
				\hline
				\multirow{4}{*}{Average} 
				&Hamming Accuracy		&87.19&81.27&85.64&88.16&87.37&88.62&\cellcolor{third}89.02&87.58&88.30&88.24&\cellcolor{second}90.47&\cellcolor{first}\textbf{90.75}\\
				&Macro-Precision 	&52.93&45.20&50.53&57.06&54.16&\cellcolor{third}60.93&57.51&55.20&57.26&55.95&\cellcolor{second}60.70&\cellcolor{first}\textbf{66.27}\\
				&Macro-Recall  	&44.91&31.23&50.30&49.77&49.71&51.94&52.20&49.22&52.31&50.94&\cellcolor{third}54.51&\cellcolor{first}\textbf{60.89}\\ 
				&Macro-F1  	&45.84&33.26&47.35&49.22&49.39&52.46&\cellcolor{third}51.85&48.09&\cellcolor{second}51.79&50.69&\cellcolor{second}54.88&\cellcolor{first}\textbf{60.29}\\
				\hline
			\end{tabular}
		}
		\caption{Multi-label Entity Recognition Performance on Ultrasound Diagnosis Tasks.}
		\label{tab:diagnosis_entity}
	\end{table*}
	\begin{table*}[!ht]
		\small 
		\centering
		\resizebox{\linewidth}{!}{%
			\begin{tabular}{c|c|cccccccccccc}
				\hline
				Anatomical	&Metric$\uparrow$&\makecell{MedGemma\\(4B)}&\makecell{LLaVA1.5\\(7B)}&\makecell{HuatuoGPT-Vision\\(7B)}& \makecell{Lingshu\\(7B)} & \makecell{LLaVA-OneVision\\(7B)}&\makecell{Qwen2-VL\\(7B)}&\makecell{LLaVA-Med\\(7B)}&\makecell{Qwen2.5-VL\\(7B)}&\makecell{LLaVA-NeXT\\(13B)}&\makecell{Gemma-3\\(12B)}&\makecell{EchoVLM\\(3B)} &EchoVLM(11B) \\  
				\hline
				\multirow{5}{*}{Breast} 
				&BLEU-1		&23.34&19.55&\cellcolor{third}28.45&\cellcolor{second}28.75&26.69&22.55&19.22&23.21&19.82&23.15&22.30&\cellcolor{first}\textbf{29.16}\\ 
				&ROUGE-1 	&30.76&26.09&\cellcolor{second}33.88&\cellcolor{third}33.69&32.76&30.36&25.91&31.13&29.35&30.97&30.14&\cellcolor{first}\textbf{33.96}\\
				&ROUGE-L  	&29.83&23.16&\cellcolor{second}32.18&\cellcolor{third}32.13&31.19&29.37&23.29&30.09&28.08&29.68&29.19&\cellcolor{first}\textbf{32.35}\\ 
				&METEOR  	&21.28&18.67&\cellcolor{third}24.55&\cellcolor{second}24.76&23.49&20.75&18.43&21.45&19.32&21.32&20.59&\cellcolor{first}\textbf{25.01}\\
				&BERTScore	&52.66&42.09&\cellcolor{first}\textbf{54.92}&\cellcolor{third}54.36&53.96&52.42&42.45&53.16&51.32&52.79&52.24&\cellcolor{second}54.58\\
				\hline
				\multirow{5}{*}{Liver} 
				&BLEU-1		&21.05&23.77&31.95&\cellcolor{second}36.79&18.91&23.64&23.73&33.33&22.76&\cellcolor{third}36.31&23.94&\cellcolor{first}\textbf{37.36}\\
				&ROUGE-1  	&28.72&27.56&46.03&\cellcolor{second}53.20&29.12&32.89&27.76&\cellcolor{third}49.28&32.38&49.13&35.74&\cellcolor{first}\textbf{54.06}\\
				&ROUGE-L  	&21.62&21.92&30.01&\cellcolor{second}32.55&21.20&24.70&21.81&31.02&23.81&\cellcolor{third}31.55&24.10&\cellcolor{first}\textbf{32.85}\\ 
				&METEOR  	&15.68&17.66&24.34&\cellcolor{second}27.33&14.62&18.63&18.03&25.18&18.62&\cellcolor{third}26.42&18.43&\cellcolor{first}\textbf{27.68}\\
				&BERTScore	&43.13&41.66&56.23&\cellcolor{second}61.23&43.76&46.23&41.25&58.72&46.05&\cellcolor{third}58.92&48.72&\cellcolor{first}\textbf{61.77}\\
				\hline
				\multirow{5}{*}{Thyroid} 
				&BLEU-1		&21.48&17.96&\cellcolor{third}27.22&\cellcolor{second}29.84&25.39&19.78&18.59&19.16&15.88&19.72&20.13&\cellcolor{first}\textbf{30.06}\\
				&ROUGE-1 	&31.55&26.51&\cellcolor{third}35.48&\cellcolor{second}37.27&33.52&31.74&27.05&30.62&28.75&31.13&31.77&\cellcolor{first}\textbf{37.37}\\
				&ROUGE-L  	&26.07&21.49&\cellcolor{third}28.77&\cellcolor{second}31.57&27.01&25.83&21.54&24.80&23.41&25.34&25.87&\cellcolor{first}\textbf{31.78}\\ 
				&METEOR  	&19.71&17.37&\cellcolor{third}23.90&\cellcolor{second}25.86&22.48&18.95&17.50&18.13&15.94&18.80&19.39&\cellcolor{first}\textbf{26.05}\\
				&BERTScore	&43.87&37.41&\cellcolor{third}47.78&\cellcolor{second}48.53&46.09&43.98&37.77&43.02&41.13&43.40&43.48&\cellcolor{first}\textbf{48.61}\\
				\hline
				\multirow{5}{*}{Average} 
				&BLEU-1		&21.96&20.43&\cellcolor{third}29.21&\cellcolor{second}31.79&23.66&21.99&20.51&25.23&19.49&26.39&22.12&\cellcolor{first}\textbf{32.19}\\
				&ROUGE-1 	&30.34&26.72&\cellcolor{third}38.46&\cellcolor{second}41.39&31.80&31.66&26.91&37.01&30.16&37.08&32.55&\cellcolor{first}\textbf{41.80}\\
				&ROUGE-L  	&25.84&22.19&\cellcolor{third}30.32&\cellcolor{second}32.08&26.47&26.63&22.21&28.64&25.10&28.86&26.39&\cellcolor{first}\textbf{32.33}\\ 
				&METEOR  	&18.89&17.90&\cellcolor{third}24.26&\cellcolor{second}25.98&20.20&19.44&17.99&21.59&17.96&22.18&19.47&\cellcolor{first}\textbf{26.25}\\
				&BERTScore	&46.55&40.39&\cellcolor{third}52.98&\cellcolor{second}54.71&47.94&47.54&40.49&51.63&46.17&51.70&48.15&\cellcolor{first}\textbf{54.99}\\
				\hline
			\end{tabular}
		}
		\caption{Out-of-Distribution Test Performance on Public Datasets Using NLP Metrics.}
		\label{tab:ood_nlp_metrics}
	\end{table*}
	\begin{table*}[!ht]
		\small 
		\centering
		\resizebox{\linewidth}{!}{%
			\begin{tabular}{c|c|cccccccccccc}
				\hline
				Anatomical	&Metric$\uparrow$&\makecell{MedGemma\\(4B)}&\makecell{LLaVA1.5\\(7B)}&\makecell{HuatuoGPT-Vision\\(7B)}& \makecell{Lingshu\\(7B)} & \makecell{LLaVA-OneVision\\(7B)}&\makecell{Qwen2-VL\\(7B)}&\makecell{LLaVA-Med\\(7B)}&\makecell{Qwen2.5-VL\\(7B)}&\makecell{LLaVA-NeXT\\(13B)}&\makecell{Gemma-3\\(12B)}&\makecell{EchoVLM\\(3B)} &\makecell{EchoVLM\\(11B)} \\  
				\hline
				\multirow{4}{*}{Breast} 
				&Hamming Accuracy		&55.36&52.89&\cellcolor{first}\textbf{57.48}&\cellcolor{third}56.98&56.55&55.09&52.64&55.58&54.78&55.72&55.09&\cellcolor{second}57.08\\ 
				&Macro-Precision 	&37.23&\cellcolor{second}56.98&36.42&36.03&36.13&36.37&\cellcolor{first}\textbf{58.13}&36.77&28.39&\cellcolor{third}45.88&28.48&36.08\\
				&Macro-Recall  	&45.83&35.21&\cellcolor{first}\textbf{51.02}&\cellcolor{third}46.77&\cellcolor{second}49.17&45.27&37.32&44.35&44.53&46.19&45.08&47.08\\ 
				&Macro-F1  	&33.09&29.40&\cellcolor{first}\textbf{38.31}&\cellcolor{second}37.66&\cellcolor{third}36.75&31.89&29.14&33.81&31.23&33.56&31.55&37.87\\
				\hline
				\multirow{4}{*}{Liver} 
				&Hamming Accuracy		&58.04&51.18&76.06&\cellcolor{second}89.61&57.04&56.32&50.73&\cellcolor{third}85.28&52.76&\cellcolor{third}85.97&63.63&\cellcolor{first}\textbf{91.06}\\
				&Macro-Precision  	&64.44&\cellcolor{second}69.26&58.43&\cellcolor{first}\textbf{68.54}&61.00&63.83&\cellcolor{third}65.78&60.85&59.23&64.16&64.31&58.57\\
				&Macro-Recall  	&26.48&20.89&45.31&\cellcolor{second}58.75&27.54&24.77&20.35&\cellcolor{third}54.00&21.53&\cellcolor{third}55.72&32.97&\cellcolor{first}\textbf{60.00}\\ 
				&Macro-F1  	&34.67&28.25&50.64&\cellcolor{second}58.91&34.84&32.50&27.36&\cellcolor{third}56.19&27.37&\cellcolor{third}57.55&40.78&\cellcolor{first}\textbf{59.25}\\
				\hline
				\multirow{4}{*}{Thyroid} 
				&Hamming Accuracy		&52.79&48.76&58.18&\cellcolor{second}62.03&55.52&52.10&49.20&49.36&46.46&51.30&53.24&\cellcolor{first}\textbf{62.40}\\
				&Macro-Precision 	&61.97&\cellcolor{second}64.22&\cellcolor{third}60.34&58.10&59.68&61.67&\cellcolor{first}\textbf{64.89}&60.39&57.66&60.18&60.31&57.77\\
				&Macro-Recall  	&32.33&25.67&40.63&\cellcolor{second}49.38&37.09&30.95&26.41&28.99&24.49&30.31&32.58&\cellcolor{first}\textbf{49.99}\\ 
				&Macro-F1  	&33.58&29.90&43.18&\cellcolor{second}49.04&38.11&31.04&30.13&26.90&19.97&30.28&33.88&\cellcolor{first}\textbf{49.08}\\
				\hline
				\multirow{4}{*}{Average} 
				&Hamming Accuracy		&55.40&50.94&63.91&\cellcolor{second}69.54&56.37&54.50&50.86&63.41&51.33&\cellcolor{third}64.33&57.32&\cellcolor{first}\textbf{70.18}\\
				&Macro-Precision 	&54.55&\cellcolor{first}\textbf{63.49}&51.73&54.22&52.27&53.96&\cellcolor{second}62.93&52.67&48.43&\cellcolor{third}56.74&51.03&50.81\\
				&Macro-Recall  	&34.88&27.26&45.65&\cellcolor{second}51.63&37.93&33.66&28.03&42.45&30.18&\cellcolor{third}44.07&36.88&\cellcolor{first}\textbf{52.36}\\ 
				&Macro-F1  	&33.78&29.18&44.04&\cellcolor{second}48.54&36.57&31.81&28.88&38.97&26.19&\cellcolor{third}40.46&35.40&\cellcolor{first}\textbf{48.73}\\
				\hline
			\end{tabular}
		}
		\caption{Out-of-Distribution Test Performance on Public Datasets Using Entity Recognition.}
		\label{tab:ood_entity_recognition}
	\end{table*}
\section{Case Study}

\begin{figure*}[t]
	\centering
	\includegraphics[width=1.0\textwidth]{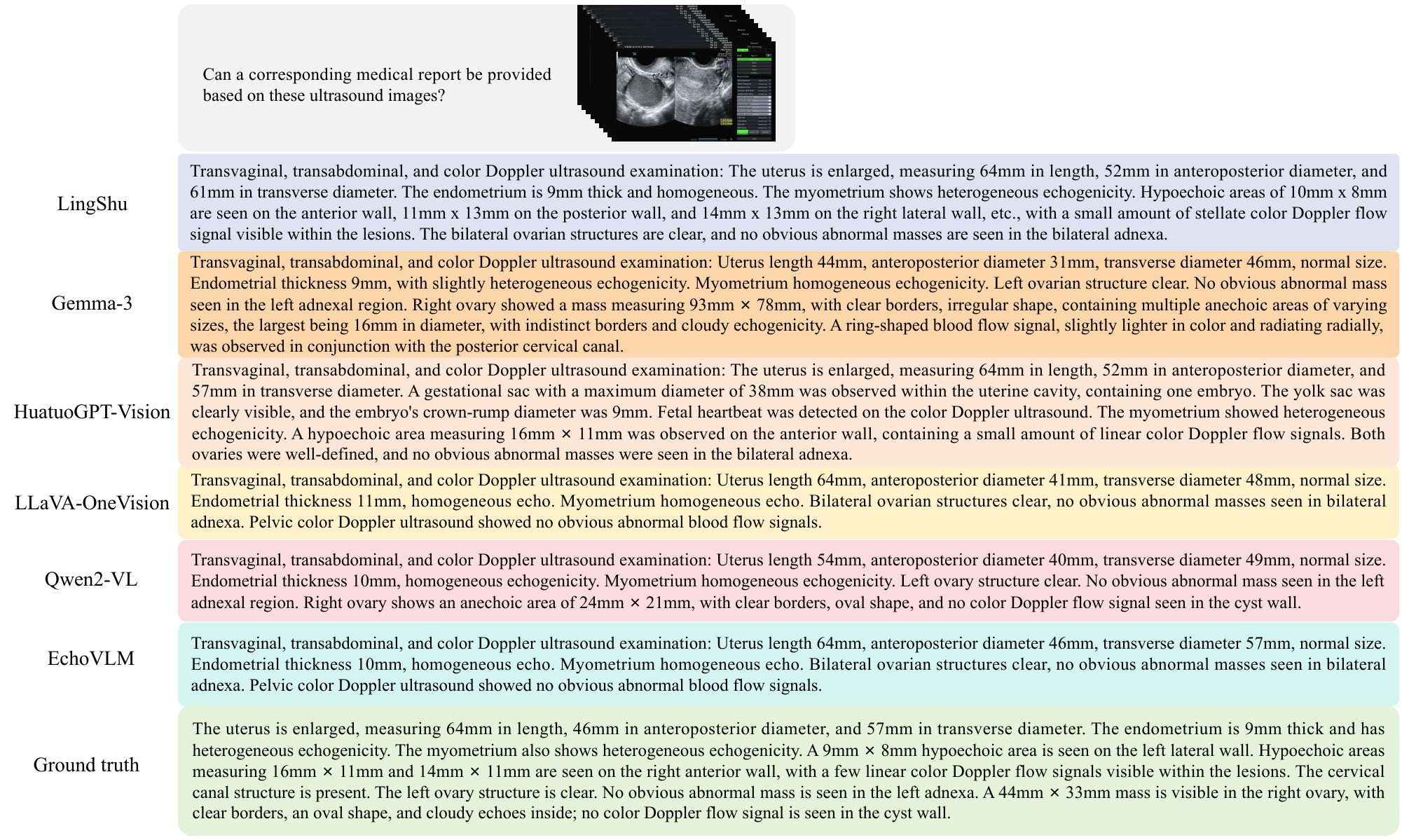}\\
	\caption{Gynecological Case Study.}
	\label{case_study1}
\end{figure*}

\begin{figure*}[t]
	\centering
	\includegraphics[width=1.0\textwidth]{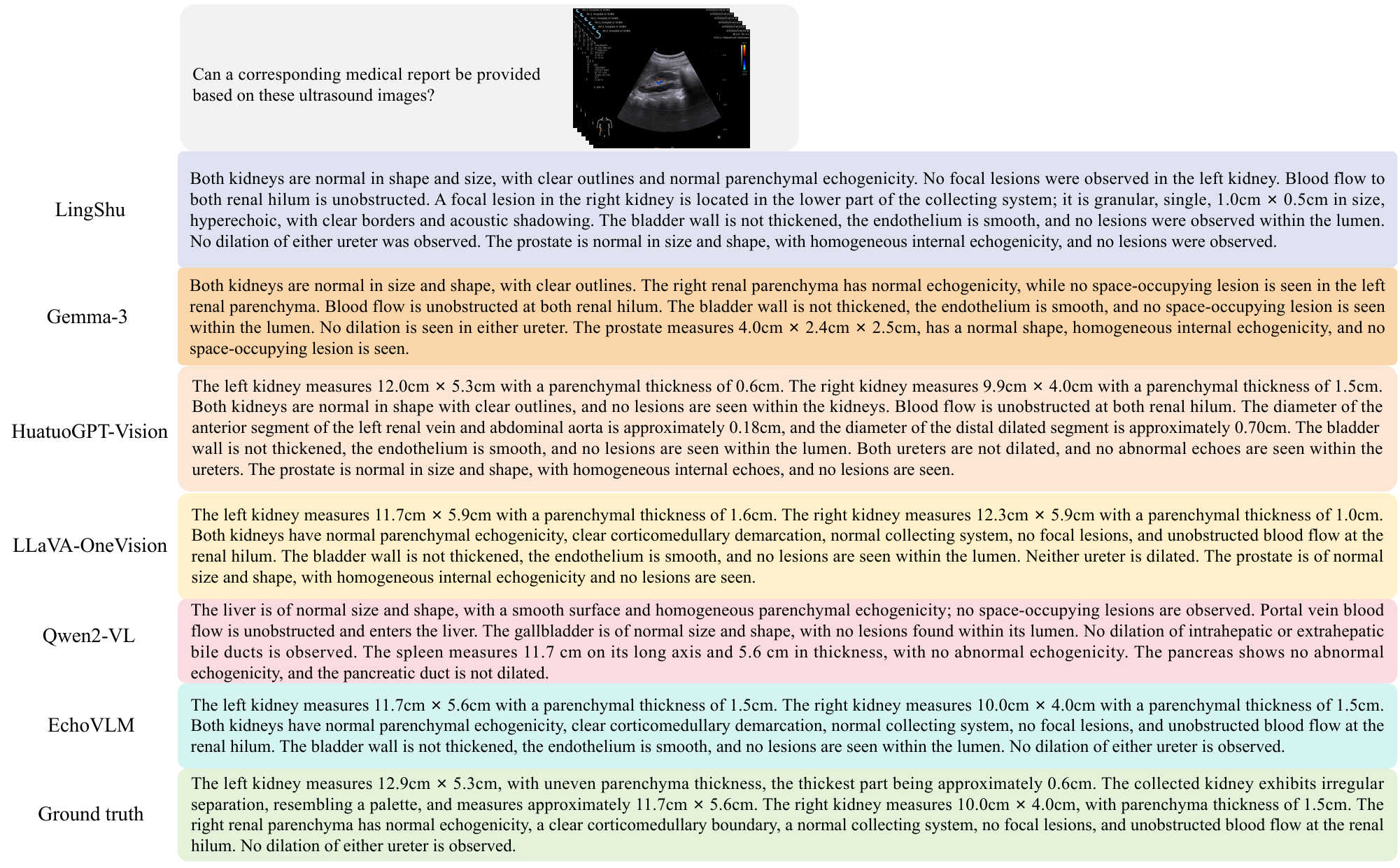}\\
	\caption{Kidney Case Study.}
	\label{case_study2}
\end{figure*}

\begin{figure*}[t]
	\centering
	\includegraphics[width=1.0\textwidth]{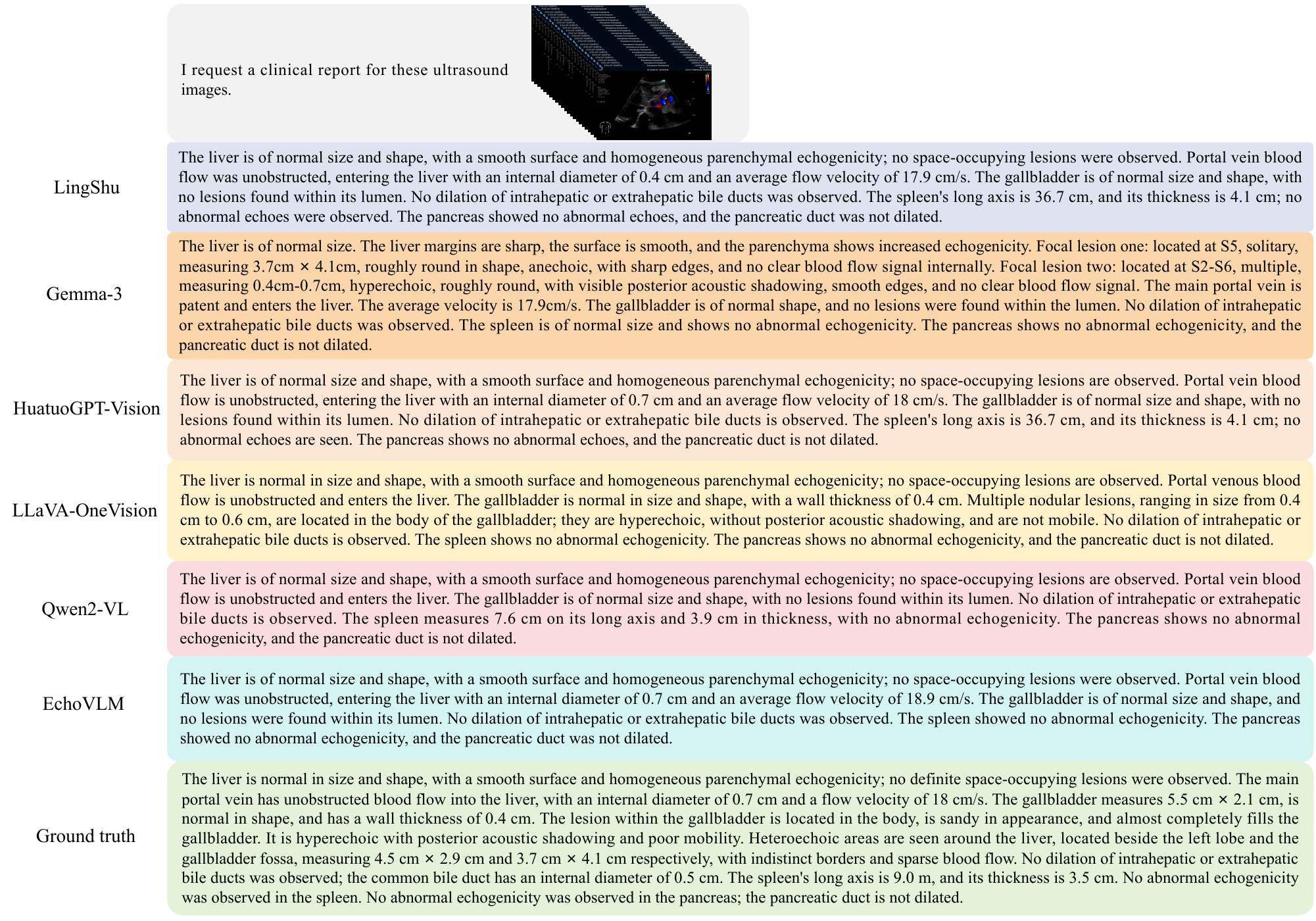}\\
	\caption{Liver Case Study.}
	\label{case_study3}
\end{figure*}

\begin{figure*}[t]
	\centering
	\includegraphics[width=1.0\textwidth]{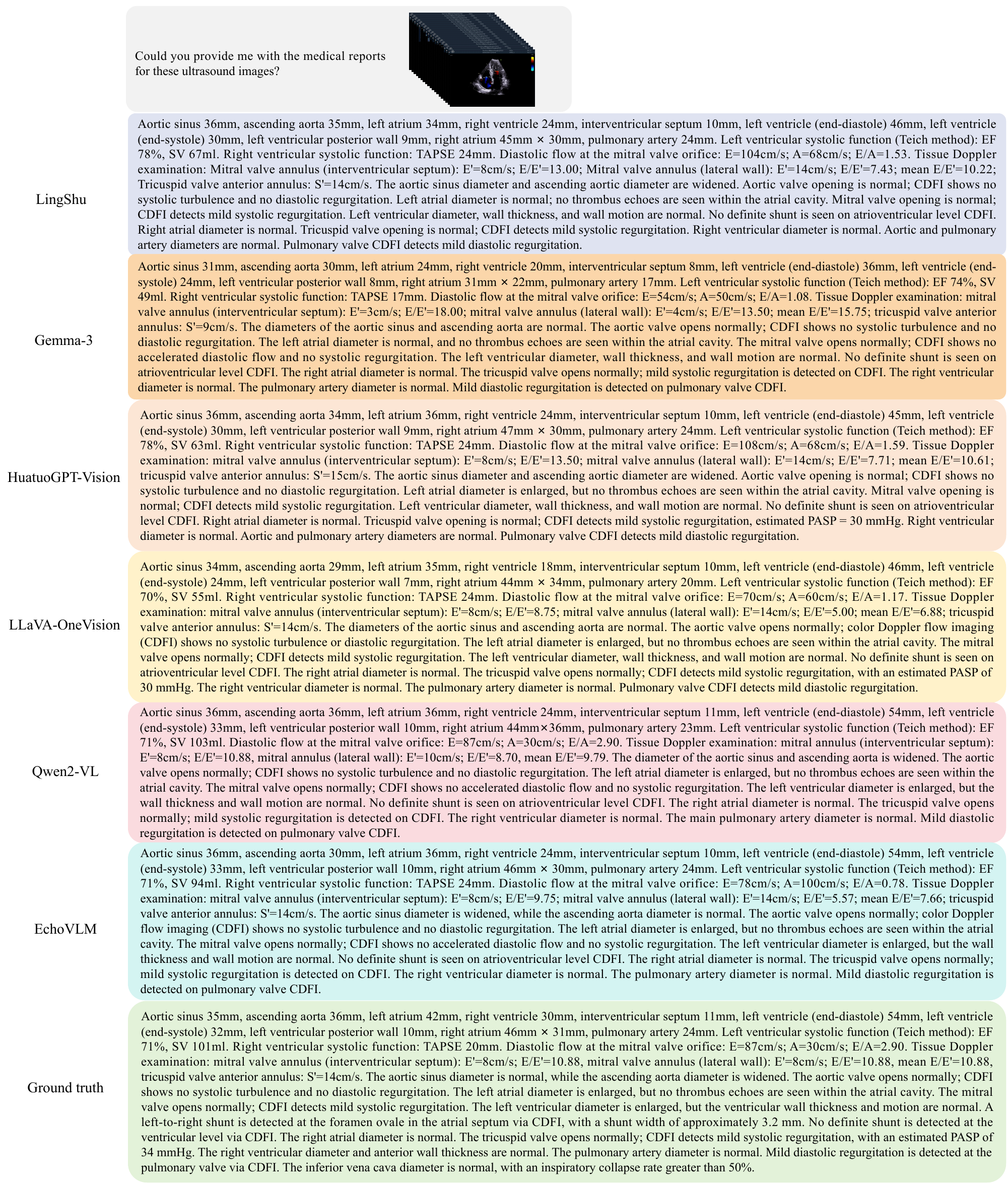}\\
	\caption{Heart Case Study.}
	\label{case_study4}
\end{figure*}

\begin{figure*}[t]
	\centering
	\includegraphics[width=1.0\textwidth]{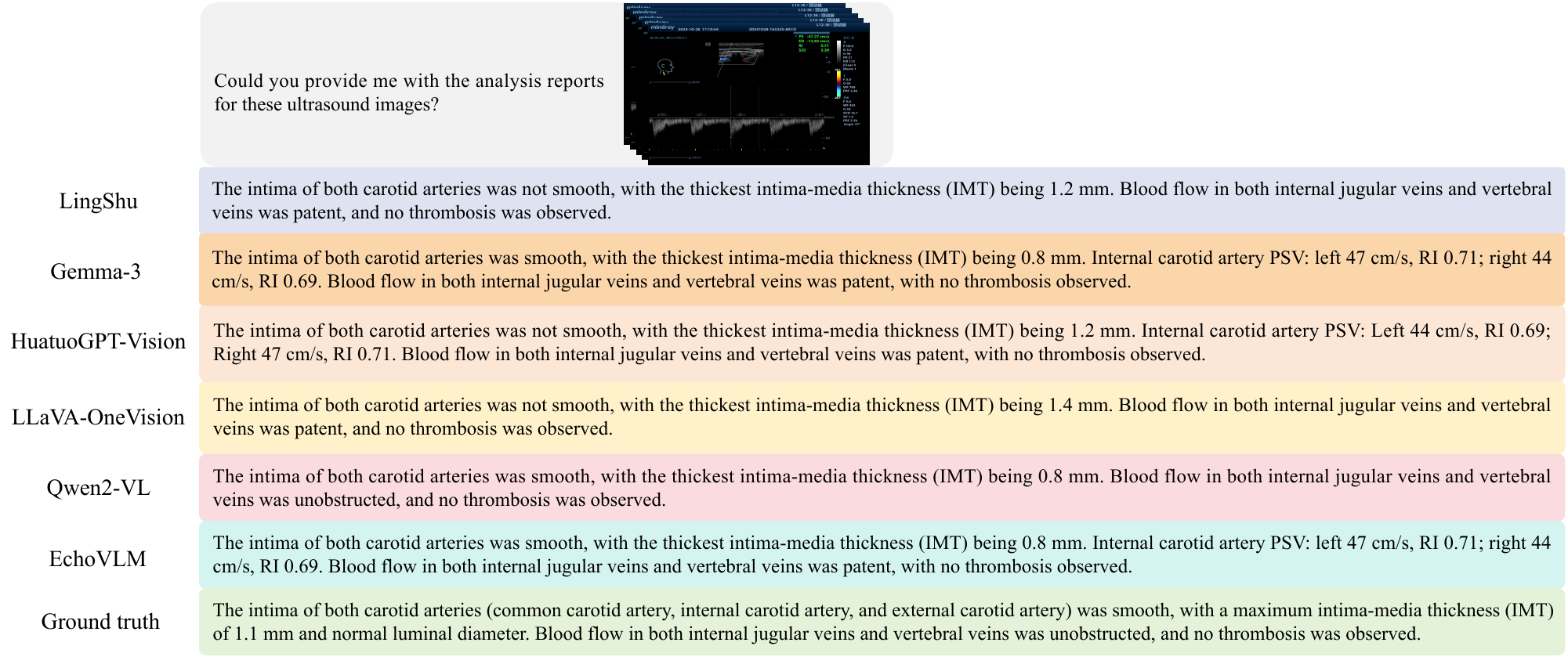}\\
	\caption{Vascular Case Study.}
	\label{case_study5}
\end{figure*}

\begin{figure*}[t]
	\centering
	\includegraphics[width=1.0\textwidth]{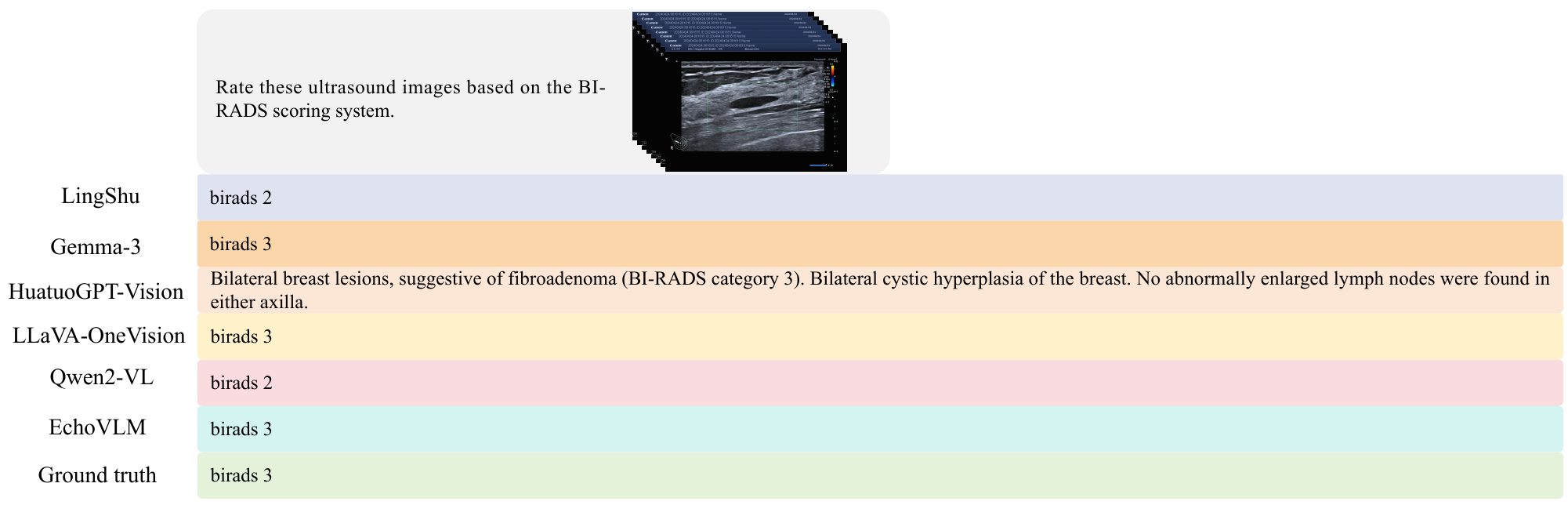}\\
	\caption{Breast Case Study.}
	\label{case_study6}
\end{figure*}

\begin{figure*}[t]
	\centering
	\includegraphics[width=1.0\textwidth]{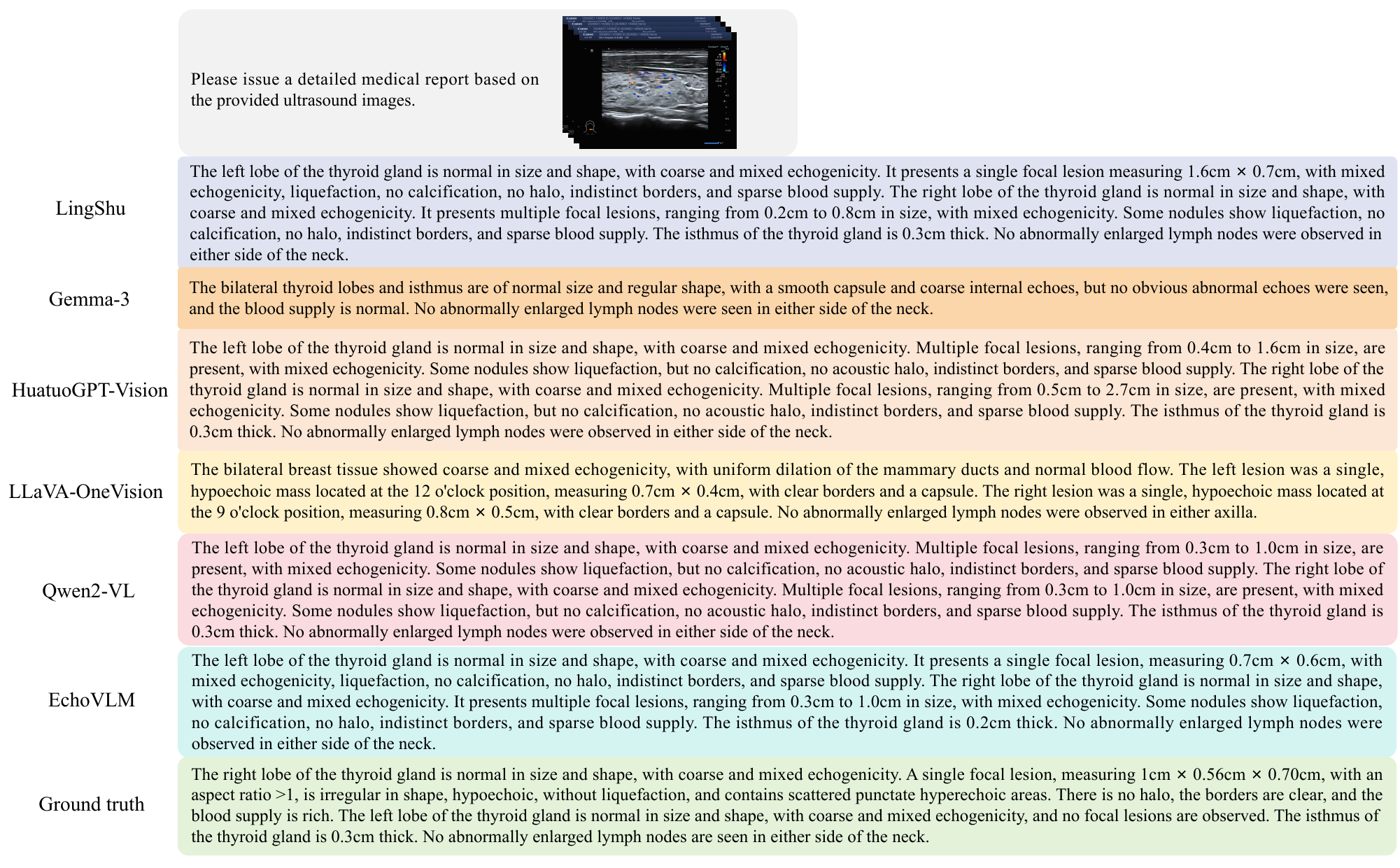}\\
	\caption{Thyroid Case Study.}
	\label{case_study7}
\end{figure*}

Figure~\ref{case_study1}--\ref{case_study7} presents an analysis of report generation cases covering different anatomical regions.
 Due to space limitations, this section randomly selects examples for discussion. This random selection method ensures an objective assessment of the model's capabilities while maintaining scientific rigor. The results show that our model exhibits significant advantages over both general-purpose and specialized ultrasound models, effectively capturing clinically meaningful information from images to support accurate reasoning and report generation. Despite these advantages, some limitations exist, including false negatives in nodule identification, suggesting that there is still room for improvement.
\section{Fine-grained Medical Entity-level Evaluation}
\label{sec:supp_entity_eval}
To provide a clinically meaningful assessment of generated medical reports, we adopt the entity-level evaluation methodology from \citet{li2024ultrasound}. This approach focuses on the accuracy of extracted medical entities rather than surface-level text similarity, which aligns better with how clinicians evaluate report quality. Table \ref{tab:report_entity} presents fine-grained entity-level performance for ultrasound report generation across seven anatomical systems, while Table \ref{tab:diagnosis_entity} shows corresponding results for ultrasound diagnosis. In both tasks, EchoVLM consistently outperforms existing multimodal models across most metrics, demonstrating its superior capability in understanding and modeling fine-grained medical entities.
\section{Out-of-Distribution (OOD) Evaluation}\label{sec:ood_evaluation}
To perform a more comprehensive assessment of the models, we conduct Out-of-Distribution (OOD) evaluation on the public ultrasound dataset introduced by \citet{li2024ultrasound}, with the primary objective of verifying the generalization capability of the evaluated models to unseen ultrasound data. Notably, none of the samples from this public dataset were included in the model training pipeline, thereby eliminating potential data leakage and ensuring the rigor, impartiality, and reliability of the experimental evaluations.
To systematically evaluate the OOD generalization capability of all models, we adopt two complementary sets of evaluation metrics, and the detailed OOD test results are presented in the subsequent tables. Table \ref{tab:ood_nlp_metrics} quantifies the OOD performance using conventional natural language processing (NLP) metrics, which assess the coherence, relevance, and accuracy of the generated text. In contrast, Table \ref{tab:ood_entity_recognition} employs fine-grained entity recognition metrics, which prioritize the precision of medical entity extraction—a critical requirement for clinical ultrasound diagnosis. Collectively, these results demonstrate the models’ adaptability to unseen ultrasound data, a property that is critical for practical clinical ultrasound diagnosis.


\end{document}